%% file: main.tex
\documentclass[10pt,twocolumn,letterpaper]{article}

\usepackage{cvpr}              % To produce the CAMERA-READY version
\input{preamble}

\definecolor{cvprblue}{rgb}{0.21,0.49,0.74}
\usepackage[pagebackref,breaklinks,colorlinks,allcolors=cvprblue]{hyperref}

\usepackage{multirow} % 用于合并表格的行
\usepackage{tikz}
\usepackage{amsthm}
\usepackage{algorithm}
\usepackage[noend]{algpseudocode}
\usepackage{amssymb}
\usepackage{amsmath}
\usepackage{algpseudocode}

\newtheorem{definition}{Definition}
\newtheorem{lemma}{Lemma}
\newtheorem{theorem}{Theorem}

\def\paperID{19010} % *** Enter the Paper ID here
\def\confName{CVPR}
\def\confYear{2026}

\title{SPEGC: Continual Test-Time Adaptation via Semantic-Prompt-Enhanced Graph Clustering for Medical Image Segmentation}

\author{Xiaogang Du,\quad Jiawei Zhang,\quad Tongfei Liu,\quad Tao Lei,\thanks{Corresponding author.}\quad Yingbo Wang\\
Shaanxi Joint Laboratory of Artificial Intelligence, Shaanxi University of Science and Technology.\\
Xi'an, China\\
\tt\small duxiaogang@sust.edu.cn, hello.jiawei@outlook.com, leitao@sust.edu.cn}
\begin{document}
\maketitle
\input{sec/0_abstract}    
\input{sec/1_intro}

\input{sec/2_relatedwork}

\input{sec/3_methods}
\input{sec/4_experiments}
\input{sec/5_conclusion}

{
    \small
    \bibliographystyle{ieeenat_fullname}
    \bibliography{main}
}

\input{sec/A1_suppl}
\input{sec/A2_algorithm}

\input{sec/A3_additional_visualization}


\end{document}

%% file: preamble.tex
\usepackage{microtype}

%% file: sec/0_abstract.tex
\begin{abstract}
In medical image segmentation tasks, the domain gap caused by the difference in data collection between training and testing data seriously hinders the deployment of pre-trained models in clinical practice. Continual Test-Time Adaptation (CTTA) aims to enable pre-trained models to adapt to continuously changing unlabeled domains, providing an effective approach to solving this problem. However, existing CTTA methods often rely on unreliable supervisory signals, igniting a self-reinforcing cycle of error accumulation that culminates in catastrophic performance degradation. To overcome these challenges, we propose a CTTA via Semantic-Prompt-Enhanced Graph Clustering (SPEGC) for medical image segmentation. First, we design a semantic prompt feature enhancement mechanism that utilizes decoupled commonality and heterogeneity prompt pools to inject global contextual information into local features, alleviating their susceptibility to noise interference under domain shift. Second, based on these enhanced features, we design a differentiable graph clustering solver. This solver reframes global edge sparsification as an optimal transport problem, allowing it to distill a raw similarity matrix into a refined and high-order structural representation in an end-to-end manner. Finally, this robust structural representation is used to guide model adaptation, ensuring predictions are consistent at a cluster-level and dynamically adjusting decision boundaries. Extensive experiments demonstrate that SPEGC outperforms other state-of-the-art CTTA methods on two medical image segmentation benchmarks. The source code is available at \url{https://github.com/Jwei-Z/SPEGC-for-MIS}.

%The ABSTRACT is to be in fully justified italicized text, at the top of the left-hand column, below the author and affiliation information.
%Use the word ``Abstract'' as the title, in 12-point Times, boldface type, centered relative to the column, initially capitalized.
%The abstract is to be in 10-point, single-spaced type.
%Leave two blank lines after the Abstract, then begin the main text.
%Look at previous \confName abstracts to get a feel for style and length.
\end{abstract}

%% file: sec/1_intro.tex
\begin{figure}[t]
  \centering

  \includegraphics[width=1\linewidth]{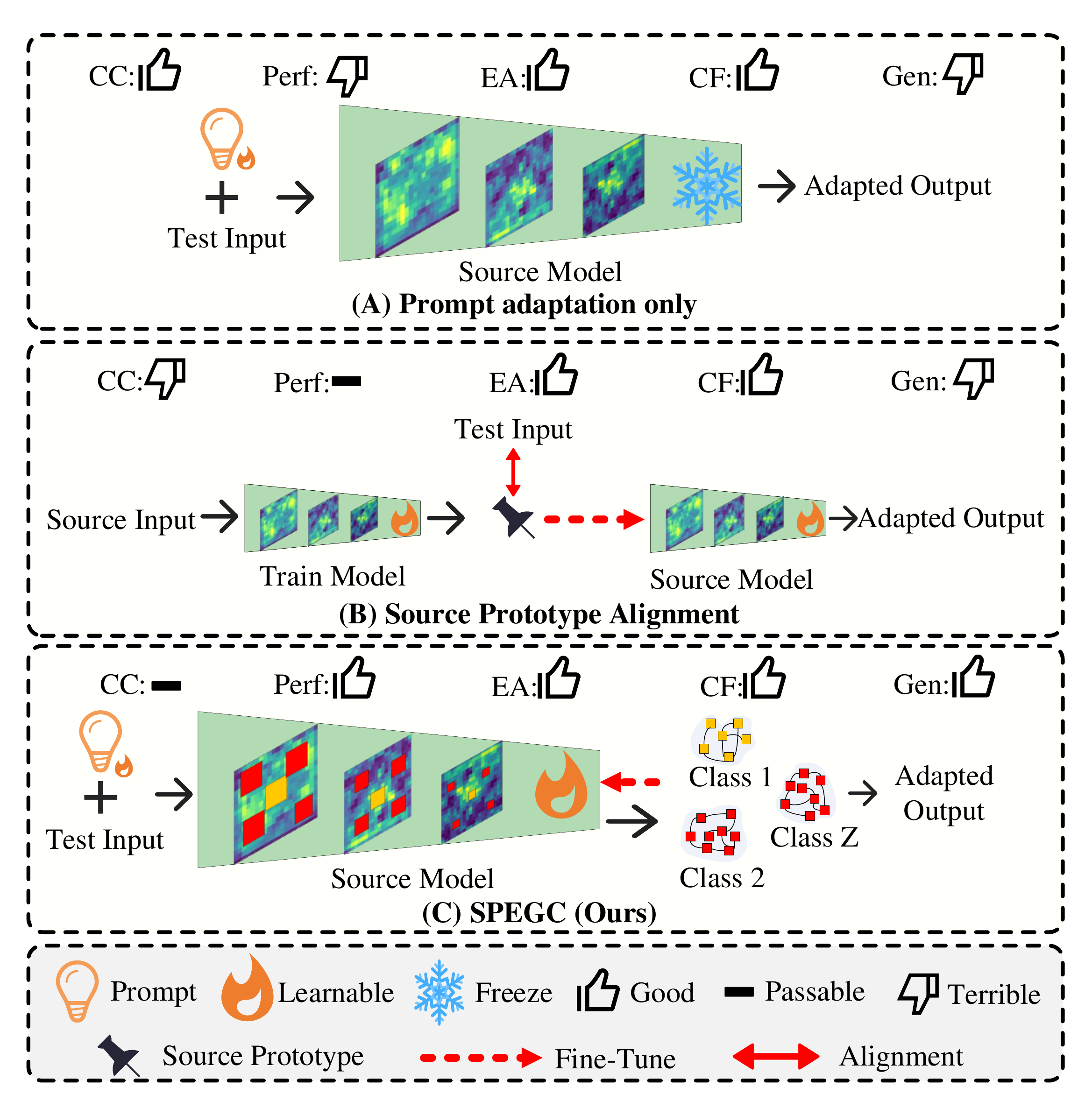}

  \caption{Conceptual comparison of different paradigms in CTTA. Compared to (A) Prompt adaptation only and (B) Source Prototype Alignment, (C) our SPEGC innovatively leverages graph clustering to extract structural information, achieving "Good" performance across Performance (Perf), Generalization (Gen), Error Accumulation (EA), and Catastrophic Forgetting (CF), while maintaining an acceptable Computational Complexity (CC).}

  \label{fig:motivation}
\end{figure}
\vspace{-3mm}
\section{Introduction}
\label{sec:intro}

Medical image segmentation is a critical tool in clinical practice, but its deployment is severely hindered by domain shift. Models that perform well on source-domain data suffer significant performance degradation when applied to target-domain data from different protocols, operators, or scanners \cite{sankaranarayanan2018learning,ghafoorian2017transfer}. While Unsupervised Domain Adaptation (UDA) \cite{chen2022reusing,wang2019boundary,yang2020fda,zuo2021attention} addresses this, it typically requires access to complete source data and large target-domain batches \cite{yang2022dltta}. These requirements are often impractical in clinical settings due to patient privacy restrictions and the sequential, single-sample arrival of test data.
Test-Time Adaptation (TTA) \cite{nguyen2023tipi,niu2022efficient,niu2023towards,sun2020test,wang2021exploring,zhang2023domainadaptor,zhang2022memo} offers a more realistic, source-free paradigm by updating the model at inference time. Popular TTA methods rely on self-supervised auxiliary losses \cite{wang2021exploring}, introduce regularizers to prevent drastic parameter changes \cite{niu2022efficient,niu2023towards}, gradient regularization \cite{chen2025gradient}, or alignment with pre-trained priors \cite{lv2025test}(See \cref{fig:motivation}(B)). However, these approaches often suffer from error accumulation and, critically, assume a static target domain. In contrast, real-world clinical data arrives as a continually evolving stream, with test samples often requiring immediate, individual predictions. Consequently, Continual Test-Time Adaptation (CTTA) \cite{wang2022continual} has emerged to handle these sequential distribution changes. This setting is far more challenging, exacerbating error accumulation \cite{cascante2021curriculum,liang2019exploring} and catastrophic forgetting \cite{niu2022efficient,lamers2023clustering}. While recent prompt-based methods \cite{gan2023decorate,chen2024each} offer a lightweight solution by learning prompts in the input space, their performance is inherently limited as the core model parameters remain frozen(See \cref{fig:motivation}(A)).

To overcome these limitations, we propose a novel CTTA framework via \textbf{S}emantic-\textbf{P}rompt-\textbf{E}nhanced \textbf{G}raph \textbf{C}lustering (SPEGC), which adapts by reasoning on high-order structural abstractions. However, in domain shift, local features of unlabeled test samples are highly susceptible to noise. SPEGC tackles this via a dual mechanism. First, we design a Semantic Prompt Feature Enhancement (SPFE) mechanism, which uses two decoupled, learnable prompt pools to retrieve and inject robust global contextual information into the noisy local node features. Second, based on these enhanced features, we design a Differentiable Graph Clustering Solver (DGCS). This solver reframes the graph partitioning task as an optimal transport problem, allowing it to distill a raw, noisy similarity matrix into a refined, high-order structural representation. This robust structural representation is then used to guide the model's adaptation in an end-to-end manner, ensuring cluster-level consistency and dynamically adjusting decision boundaries (See \cref{fig:motivation}(C)).
Based on the above, our method mainly has the following advantages: \textbf{(1) Robustness to feature-level noise}. By injecting decoupled global contextual information via the semantic prompts. This makes the adaptation significantly less susceptible to the interference and style variations. \textbf{(2) Explicit preservation of key semantics}. The commonality prompt pool, guided by a dedicated clustering loss, establishes a stable semantic anchor. This mechanism explicitly preserves shared and cross-domain knowledge, effectively mitigating catastrophic forgetting during continual adaptation introduced by domain shifts. \textbf{(3) Stable and high-order supervision}. Our method generates its supervisory signal from a refined, high-order structural representation distilled by the DGCS. This graph-based guidance ensures adaptation is driven by the inherent structure of the data, which is more reliable and robust against error accumulation.

Our contributions can be summarized as follows.
\begin{itemize}
\item We propose SPEGC, a novel CTTA framework that leverages high-order structural information via differentiable graph clustering to guide self-regulation in unseen domains.

\item We design SPFE with decoupled prompt pools to enhance local node features with robust global context, making them resilient to domain shifts.

\item We design DGCS, based on optimal transport, to distill a refined, structurally-consistent edge similarity matrix in an end-to-end manner.

\item Extensive experiments on two medical image segmentation benchmarks show SPEGC achieves state-of-the-art performance in  single-source settings.
\end{itemize}

%% file: sec/2_relatedwork.tex
\section{Related Work}

\subsection{Clustering-based Image Segmentation}
Clustering-based image segmentation has a long history. Before the renaissance of deep learning, traditional methods \cite{ray1999determination} relied on low-level features such as texture or color, which limited their ability to capture high-level semantics \cite{yu2022k}. Recent research has shifted towards utilizing convolutional neural networks to extract features \cite{long2015fully,liang2022gmmseg,zhou2022rethinking,wang2021exploring,li2022deep,liang2023logic,li2023logicseg,cheng2023segment,li2023semantic,li2024omg}, clustering pixels into semantic regions as a post-processing step \cite{neven2019instance,kong2018recurrent,feng2023clustering,yin2022proposalcontrast,quan2024clustering,feng2024interpretable3d}. Furthermore, research has also begun to shift towards query-based Transformer methods \cite{cheng2022masked}. For example, Yu et al. \cite{yu2022cmt,yu2022k} revisited the relationship between pixel features and object queries by reformulating cross-attention as a clustering solver. Building on this, Liang et al. \cite{liang2023clustseg} introduced a recurrent cross-attention mechanism, unlocking the potential of iterative clustering for pixel grouping. Ding et al. \cite{ding2024clustering} further extended the clustering mechanism from 2D images to 3D volumetric data and establishes connections between slices of the 2D network.

However, these methods primarily serve segmentation tasks within static domains, and their clustering process is either a post-processing step or designed for static grouping. They lack the ability to leverage the dynamically changing graph structural information within the data stream to guide model adaptation. When confronted with domain shift, the similarity matrices computed directly from features are often fraught with noise, and existing methods cannot explicitly refine a structure that characterizes semantic consistency to supervise the model during optimization. In contrast, our SPEGC constructs a DGCS to distill a refined edge similarity matrix in an end-to-end manner. This enables us to directly leverage the high-order structural information inherent in the data to guide the model's self-regulation on unseen domains.

\subsection{Continual Test-Time Adaptation}

TTA aims to adapt a pre-trained model from the source domain to test data during the inference stage, in a source-free and online manner \cite{sun2020test,wang2021exploring}. Popular methods update the model by constructing self-supervised auxiliary tasks \cite{chen2022contrastive,gandelsman2022test,nguyen2023tipi,zhang2022memo,zhang2024testfit}, adjusting Batch Normalization (BN) statistics \cite{mirza2022norm,wang2023dynamically}, or filtering source-friendly targets \cite{wu2026sictta}. Recently, some studies have begun exploring more complex structural information to enhance the robustness of TTA. Lv et al. \cite{lv2025test} constructs graph structures to characterize these priors as a stable reference for cross-domain alignment. However, these methods ignore the continuously changing target domains found in most real-world scenarios [96].

Recently, CTTA \cite{wang2022continual} was proposed to address the challenge of continuously changing data distributions. Although many methods \cite{dobler2023robust,niu2022efficient,niu2023towards,song2023ecotta,wang2022continual,yang2022dltta,zhang2023domainadaptor} attempt to mitigate these problems through model resetting, regularization, or optimizing self-supervised losses, they predominantly rely on pixel-level or instance-level supervisory signals. Another class of CTTA methods \cite{chen2024each,liu2025efficient} turns to freezing the backbone network and learning lightweight visual prompts in the input space, aligning the BN layer statistics by updating only the prompts. Input-level consistency, whether guided by pixel-level signals or BN layer statistical alignment, remains fragile when faced with severe domain shifts. They lack a mechanism to explicitly capture and enhance local features that are less robust to noise and style variations. In contrast, we propose a feature enhancement module based on semantic prompts. This module utilizes a commonality prompt pool and a heterogeneity prompt pool to dynamically retrieve and decouple global information via reverse-attention and attention mechanisms, thereby generating enhanced features. This enables SPEGC to efficiently adapt to new domains while retaining memory of core semantics, effectively mitigating catastrophic forgetting.

%% file: sec/3_methods.tex
\section{Method}
\label{sec:formatting}

\subsection{Overall Structure}

\begin{figure*}[t]
  \centering
  \includegraphics[width=1\linewidth]{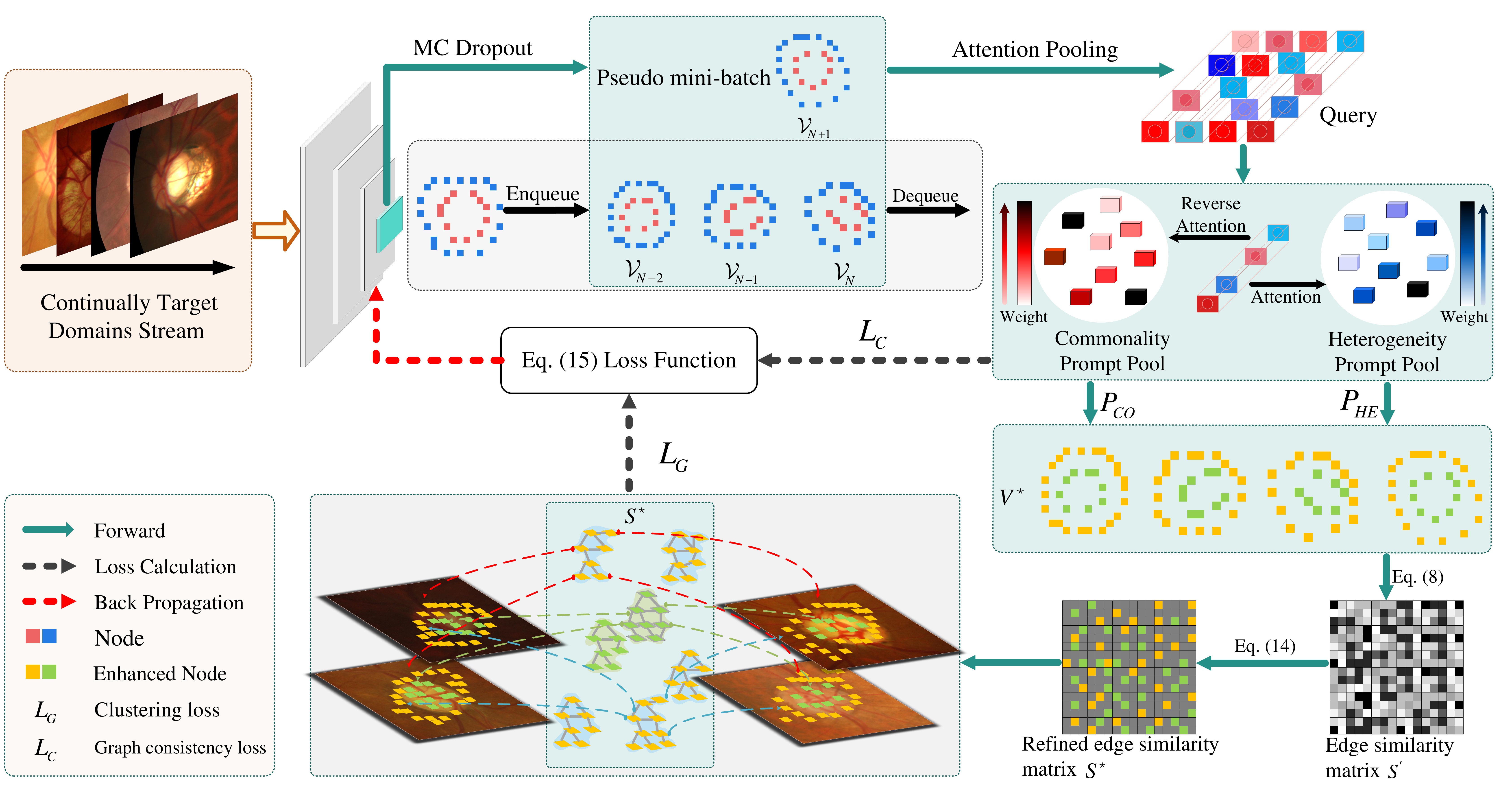}
  
  \caption{Overview of the SPEGC. For the continual target domains stream, we first extract local node features ($V$) and construct a Pseudo mini-batch via a feature queue (Enqueue/Dequeue). Subsequently, the Semantic Prompt Feature Enhancement (SPFE, \cref{sec:SPFE}) module utilizes Attention Pooling to generate a Query, retrieving information from the decoupled commonality ($P_{CO}$) and heterogeneity ($P_{HE}$) prompt pools to generate the enhanced batch features $V^{\star}$, thereby mitigating noisy information from the domain shift. Based on these features, the Differentiable Graph Clustering Solver (DGCS, \cref{sec:DGCS}) computes the initial edge similarity matrix $S^{\prime}$ and refines it end-to-end into $S^{\star}$. Finally, the segmentation network is fine-tuned via backpropagation to achieve efficient adaptation (\cref{sec:LF}).
  }
  
  \label{fig:Structure}
\end{figure*}

In CTTA, we first pre-train the model $f$ using data from the source domain $D_S = (Y_S, X_S)$. Subsequently, we adapt $f$ to dynamic target domains, denoted as $D_1 , D_2 , ..., D_T$. Inspired by \cite{ding2024clustering,liu2025efficient}, We propose SPEGC, an adaptation pipeline jointly driven by semantic prompt enhancement and differentiable graph clustering. The core idea of SPEGC is to first enhance noisy, domain-shifted image features using semantic prompts and then distill these enhanced features into a differentiable optimization objective via our DGCS. This objective, representing the intrinsic cluster-level organization of the data, guides self-regulation and adaptation of the model to unseen domains, as illustrated in \cref{fig:Structure}.

SPEGC comprises two core components: SPFE and DGCS. First, SPFE leverages decoupled prompt pools to inject global contextual information into local node features, yielding robustly enhanced features. Second, DGCS distills the resulting global similarity matrix into a refined edge similarity matrix that captures the intrinsic semantic clustering structure. Finally, SPEGC is jointly optimized via a graph consistency loss and a clustering loss.

\subsection{Semantic Prompt Feature Enhancement}
\label{sec:SPFE}

Given an input image $x_i$, we first extract visual features \cite{li2023sigma++} using a ResNet \cite{he2016deep} backbone. As model predictions can become unreliable under domain shift, we quantify this instability by spatially estimating the uncertainty of the prediction via MC Dropout \cite{gal2016dropout}. Specifically, we perform $t$ stochastic forward passes on $x_i$ to generate $t$ distinct deep feature maps $\{F_1, F_2, \dots, F_t\}$. The variance of these feature outputs serves as the uncertainty metric $U$ at each spatial location $(j, k)$:

\vspace{-5mm}

\begin{equation}
  U(j,k)=\frac{1}{t}\sum_{i=1}^{t}\left|\left|F_i(j,k)-\bar{F}(j,k)\right|\right|_2^2.
  \label{eq:1}
\end{equation}
Where $F_i(j,k)$ is the feature vector at position $(j,k)$ from the $i$-th forward pass, and $\bar{F}(j,k)$ is the average feature vector at that position over the $t$ passes.

To construct a high-quality graph, we leverage the uncertainty map $U(j,k)$ to identify the $p\%$ of foreground nodes exhibiting the lowest uncertainty. From this set of stable features, we spatially uniformly sample \cite{li2022sigma} $n_i$ nodes. These nodes are then mapped to an $h$-dimensional graph space via a non-linear projection, generating the node feature matrix $V_i \in \mathbb{R}^{n_i \times h}$.The features within $V_i$ are inherently local, lacking global context. This locality makes them highly susceptible to noise and style variations under domain shift, which consequently degrades the quality of the downstream similarity matrix $S$. To mitigate this vulnerability, we design a semantic prompt-based feature enhancement strategy to inject global decoupled contextual information into these local node features. The strategy proceeds as follows:

We first aggregate $V_i$ into a global context query $\hat{q_i} \in \mathbb{R}^{h}$. To avoid the loss of information of simple average pooling, we employ an attention-based dynamic pooling mechanism using a learnable vector $c_{p} \in \mathbb{R}^{h}$:
\begin{equation}
\hat{q_i}=V_i^T\text{Softmax}(V_i c_{p}).
\label{eq:2}
\end{equation}

The query $\hat{q_i}$, derived from the input, inherently encodes instance-specific semantics. Consequently, the retrieval mechanisms are strategically designed to either align with or diverge from this specificity. To isolate non-shared, domain-specific information (HE), a standard attention mechanism \cite{vaswani2017attention} computes weights $\alpha_{HE}^i \in \mathbb{R}^{M}$ to identify features in $P_{HE}$ that best match $\hat{q_i}$. This process employs Softmax to amplify the most relevant discriminative signals, ensuring retrieval is sharply focused on class-separating patterns. Conversely, to capture shared, cross-domain semantics (CO), a reverse-attention mechanism computes weights $\alpha_{CO}^i \in \mathbb{R}^{M}$ to identify features in $P_{CO}$ that mismatch the instance-specific $\hat{q_i}$. This mechanism utilizes ReLU to effectively implement the truncation of matching scores and selectively activate only the commonality features:

\vspace{-5mm}

\begin{align}
  \alpha_{CO}^i = ReLU(-\gamma(\hat{q}_i, P_{CO}))
  \label{eq:3}\\
  \alpha_{HE}^i = Softmax(\gamma(\hat{q}_i, P_{HE}))
  \label{eq:4}
\end{align}
Where $\gamma$ denotes cosine similarity computed between the query and each prompt in the respective pool. The resulting commonality prompt $p_{CO}(i)$ and heterogeneity prompt $p_{HE}(i)$ are the weighted sums of their respective pools:

\begin{align}
  p_{CO}(i) &= \sum_{j=1}^M \alpha_{CO,j}^i \cdot P_{CO,j}, \label{eq:5} \\
  p_{HE}(i) &= \sum_{j=1}^M \alpha_{HE,j}^i \cdot P_{HE,j}. \label{eq:6}
\end{align}

These prompts are then applied as decoupled context biases to the original node features $V_i$ to yield the final enhanced features $\mathbf{V}^*_i$:

\begin{equation}
     \mathbf{V}^*_i=V_i+p_{CO}(i)+p_{HE}(i).
     \label{eq:7} 
\end{equation}

\subsection{Differentiable Graph Clustering Solver}
\label{sec:DGCS}

To capture broader structural relationships, we maintain a queue of enhanced node features from the $N$ previous images. This queue is combined with the enhanced features $\mathbf{V}^*_i$ of the current image $x_i$ to form an aggregated pseudo-batch, $\mathbf{V}^* = \{{ \mathbf{V}^*_i \in \mathbb{R}^{n_i \times h} \}}_{i=1}^{B}$, where $B=N+1$. From this batch, which contains a total of $V = \sum_{i=1}^{B} n_i$ nodes, we compute the global similarity matrix $S \in \mathbb{R}^{V \times V}$ using two lightweight and learnable projections, $W_q \in \mathbb{R}^{h \times h}, W_k \in \mathbb{R}^{h \times h}$:

\begin{equation}
    S = \frac{(\mathbf{V}^* W_q) (\mathbf{V}^* W_k)^{\text{T}}}{\sqrt{h}}.
    \label{eq:eq8}
\end{equation}

Note that we omit Softmax normalization in \cref{eq:eq8}. Applying Softmax would force each node's affinities into a probability distribution, thereby diluting the high-confidence signal from strongly correlated nodes.
A key challenge lies in distilling this raw, noise-laden similarity matrix $S$ into a refined graph that faithfully represents the semantic clustering structure. We design DGCS to address this. DGCS reformulates the discrete combinatorial problem of selecting $Z$ optimal clusters into a continuous and differentiable optimization problem, where $Z$ is a hyperparameter.

A robust cluster should be composed of a high-density center and a set of low-density neighbors gravitating towards it, ultimately forming $Z$ most representative foreground clusters. First, we define the density $D(v_i)$ of a node $v_i$:
\begin{equation}
     D(v_i) = \sum_{j=1}^{r} S^+(i,j).
     \label{eq:9} 
\end{equation}
Where $S^+=\text{ReLU}(S)$ is used to remove negative correlations. Based on node densities, we define the edge similarity matrix $S^{\prime}$:
\begin{equation}
     S^{\prime}(i,j) = \text{ReLU}(S(i,j)) \cdot \sigma\left( \frac{D(x_j) - D(x_i)}{\tau} \right).
     \label{eq:10} 
\end{equation}
Where $\sigma$ represents the Sigmoid function and $\tau$ is a temperature parameter.

Motivated by the property that a spanning forest with $Z$ connected components contains exactly $k = V - Z$ edges, we use $k$ to establish a principled global sparsity budget. Rather than enforcing strict acyclicity or discrete forest constraints, we leverage this budget to reformulate the task as a differentiable global $k$-edge sparsification problem, selecting soft edge probabilities from the possibilities $E=V^{2}$ in $S^{\prime}$.

We first flatten the directed affinity matrix $S^{\prime}$ into an edge affinity vector $d \in \mathbb{R}^{E}$. Referring to \cite{wang2023deep}, we frame the selection as a binary assignment by constructing a cost matrix $D \in \mathbb{R}^{E \times 2}$. The first column $D_{:,1}$ represents the cost of rejecting an edge, while the second column $D_{:,2}$ represents the cost of selecting it. Let $d_{max}=\max(d)$ and $d_{min}=\min(d)$, the costs are:

\begin{equation}
     D_{i,1} = d_{max} - d_i,
     \label{eq:11} 
\end{equation}
\begin{equation}
     D_{i,2} = d_{min} - d_i.
     \label{eq:12} 
\end{equation}

Our goal is to find the optimal, entropy-regularized transport plan $\Gamma^*$ \cite{cuturi2013sinkhorn} that minimizes the total cost:

\begin{equation} 
\label{eq:13}
\begin{aligned}
    \Gamma^* = & \arg \min_{\Gamma} \langle \Gamma, D \rangle + \theta h(\Gamma) \\
    & \text{s.t. } \Gamma \mathbf{1}_2 = r, \quad \Gamma^T \mathbf{1}_E = c.
\end{aligned}
\end{equation}
Where $\langle \cdot, \cdot \rangle$ is the inner product of Frobenius and $h(\Gamma) = -\sum_{i,j}\Gamma_{i,j} \log(\Gamma_{i,j})$ is the entropy regularization. The marginal constraints are critical: $r=\mathbf{1}_E \in \mathbb{R}^E$ enforces that every edge must be assigned, and $c=[E-k, k]^T \in \mathbb{R}^2$ enforces that exactly $k$ edges are selected and $E-k$ are rejected.

This problem is solved efficiently using the parallel Sinkhorn algorithm \cite{sinkhorn1964relationship}. Initializing with $\Gamma^{(0)}=\exp(-D/\theta)$ ($\theta=0.05$), the algorithm iterates the following updates until the marginals converge to $r$ and $c$:

\begin{equation} \label{eq:14}
\begin{aligned}
    \hat{\Gamma}^{(t)} & = \text{Diag}(\Gamma^{(t-1)} \mathbf{1}_2 \oslash r)^{-1} \Gamma^{(t-1)}, \\
    \Gamma^{(t)} & = \hat{\Gamma}^{(t)} \text{Diag}((\hat{\Gamma}^{(t)})^T \mathbf{1}_E \oslash c)^{-1}.
\end{aligned}
\end{equation}
Where $\oslash$ denotes element-wise division and $\text{Diag}(\cdot)$ constructs a diagonal matrix.

Upon convergence, the second column of the optimal transport plan, $\Gamma_{:,2}^{*} \in \mathbb{R}^{E}$, represents the soft probability for each of the $E$ edges being selected. This vector is then reshaped into the final refined edge similarity matrix $S^{\star}$ with $V \times V$. $S^{\star}$ approximates global consistency. For a complete mathematical proof, please refer to the supplementary materials.

\subsection{Loss Function}
\label{sec:LF}

The total loss $L$ for SPEGC is a joint objective comprising two key components: a graph consistency loss $L_G$ and a clustering loss $L_C$:

\vspace{-3mm}

\begin{equation}
    L = L_{G}+\lambda L_{C}.
    \label{eq:15} 
\end{equation}
Where $\lambda$ is a balancing hyperparameter. These components jointly drive the end-to-end fine-tuning of all model parameters at test time. Guided by the introduced prompts and graph structure, this adaptive process is effectively steered, mitigating catastrophic forgetting while efficiently adapting to unseen domain features.

\noindent\textbf{Graph consistency loss.} The graph consistency loss $L_G$ leverages the refined edge similarity matrix $S_{edge}^{F}$ to guide and calibrate the model's semantic predictions. It operates on the principle that if two nodes $v_i$ and $v_j$ are structurally similar, their corresponding semantic predictions $P_i$ and $P_j$ must be consistent. We formulate this by minimizing the KL divergence for these weighted node pairs:
\begin{equation}
    L_G = \sum_{i=1}^{V}\sum_{j=1}^{V}S_{ij}^{\star} \cdot D_{KL}(P_{j} || sg(P_{i})).
    \label{eq:16}
\end{equation}
Where $D_{KL}$ is the KL divergence and $sg(\cdot)$ is the stop-gradient operation.

\noindent\textbf{Clustering loss.} In essence, $L_{G}$ is a purely structural loss. It cannot perceive the semantic roles of the commonality prompt $p_{CO}$ and the heterogeneity prompt $p_{HE}$, nor can it guarantee functional decoupling between them. Therefore, we introduce $L_{C}$ to explicitly constrain the commonality prompt pool. Consecutive test data should share the core discriminative semantic content. $L_{C}$ forces the model to learn this commonality across different domains by compelling the commonality prompts $p_{CO}$, generated from all images in the batch, to move closer to one another in the semantic space. $L_{C}$ is computed as follows:
\begin{equation}
    L_{C} = \frac{1}{|B|} \sum_{i,j \in B} \left(1 - \frac{p_{CO}(i) \cdot p_{CO}(j)}{|p_{CO}(i)| \cdot |p_{CO}(j)|}\right).
    \label{eq:17} 
\end{equation}

%% file: sec/4_experiments.tex
\section{Experiments}
\subsection{Datasets}
\noindent\textbf{Retinal fundus segmentation task.} We employed five public datasets originating from different medical centers, all containing consistent annotations for the Optic Disc (OD) and Optic Cup (OC). These include: Domain A (RIM-ONE \cite{fumero2011rim}), Domain B (REFUGE \cite{orlando2020refuge}), Domain C (ORIGA \cite{zhang2010origa}), Domain D (REFUGE-Test \cite{orlando2020refuge}) and Domain E (Drishti-GS \cite{sivaswamy2014drishti}). During the data preprocessing stage, we followed the established protocol in \cite{chen2024each,liu2022single}, first cropping the Region of Interest (ROI) of each image to $800\times800$ pixels and subsequently applying min-max normalization.

\noindent\textbf{Polyp segmentation task.} We selected four public datasets from different centers: Domain A (BKAI-IGH-NEOPolyp \cite{ngoc2021neounet}), Domain B (CVC-ClinicDB/CVC-612 \cite{bernal2015wm}), Domain C (ETIS \cite{silva2014toward}), and Domain D (Kvasir \cite{jha2019kvasir}). Data preprocessing steps were performed according to \cite{chen2024each}, resizing all images uniformly to $800\times800$ pixels, and normalizing them using statistics derived from ImageNet.

\vspace{-1mm}
\subsection{Experiment Setup}

\noindent\textbf{Source model training.} To ensure a fair comparison, all methods adhere to the protocol from \cite{chen2024each}. Each source dataset is randomly divided into an 8:2 train/test split. All methods uniformly use a ResNet-50 \cite{he2016deep} backbone pre-trained on ImageNet. During this source training phase, we employ an SGD optimizer with a momentum of 0.9, a learning rate of 0.001, and a batch size of 8.

\noindent\textbf{CTTA setup.} For the CTTA phase, we maintain an identical experimental setup for all competing methods. The protocol simulates a real-world online scenario in which the data arrives as a stream. For each incoming test sample, the model performs a single adaptation iteration without access to any label information. The adaptation learning rate is task-specific: 0.005 for retinal fundus segmentation and 0.01 for polyp segmentation.For our method, we set the loss balance coefficient $\lambda=0.2$ and the uncertainty sampling rate $P=0.5$ (sampling nodes with the lowest 50\% uncertainty). The feature pool size is 3 (forming a pseudo-batch of size 4 with the current instance), with $Z=48$ clustering categories, $t=4$ MC Dropout passes, and $M=8$ prompts. All experiments are implemented in PyTorch and run on a single NVIDIA 3090 GPU.

\begin{table*}[t]
  \centering

  \captionsetup{justification=raggedright}
  \caption{
  Average performance (Mean ± Std.) of our SPEGC, "No adapt", and six SOTA methods on the OD/OC segmentation task, based on five experimental runs. "Domain A" denotes training on Domain A and testing on Domain B-E, and so forth for the other domains. The best results are highlighted in \textbf{\textcolor{red}{bold red}}.
  }
  \label{tab:ODOC}

  \resizebox{\textwidth}{!}{%

    \begin{tabular}{c| ccc| ccc| ccc| ccc| ccc| ccc}
      \toprule

      \multirow{2}{*}{Methods} & 

      \multicolumn{3}{c}{Domain A} & 
      \multicolumn{3}{c}{Domain B} & 
      \multicolumn{3}{c}{Domain C} & 
      \multicolumn{3}{c}{Domain D} & 
      \multicolumn{3}{c}{Domain E} & 

      \multicolumn{3}{c}{Average} \\

      \cmidrule(lr){2-4} \cmidrule(lr){5-7} \cmidrule(lr){8-10} \cmidrule(lr){11-13} \cmidrule(lr){14-16} \cmidrule(lr){17-19}

      & DSC & $E_\phi^{\max}$ & $S_{\alpha}$ 
      & DSC & $E_\phi^{\max}$ & $S_{\alpha}$ 
      & DSC & $E_\phi^{\max}$ & $S_{\alpha}$ 
      & DSC & $E_\phi^{\max}$ & $S_{\alpha}$ 
      & DSC & $E_\phi^{\max}$ & $S_{\alpha}$ 
      & DSC $\uparrow$ & $E_{\text{max}}^{b} \uparrow$ & $S_{\alpha} \uparrow$ \\
      
      \midrule \midrule

        No Adapt (ResUNet-50) \cite{diakogiannis2020resunet} & 71.92 & 88.67 & 80.84 & 79.31 & 90.42 & 84.82 & 75.42 & 89.24 & 81.62 & 63.77 & 85.49 & 78.28 & 73.32 & 89.67 & 81.81 & 72.75 & 88.70 & 81.47 \\
        
        \midrule
        
        SAR(ICLR'23) \cite{niu2023towards} & 74.23$\pm$6.13 & 90.17$\pm$0.34 & 83.59$\pm$0.19 & 80.23$\pm$2.54 & 93.06$\pm$0.43 & 85.89$\pm$0.17 & 72.22$\pm$4.97 & 92.15$\pm$0.16 & 83.41$\pm$0.18 & 70.19$\pm$1.22 & 86.31$\pm$0.17 & 79.47$\pm$0.78 & 70.35$\pm$5.31 & 88.53$\pm$0.87 & 86.70$\pm$0.30 & 73.44 & 90.14 & 83.81 \\

        Domain Adaptor(CVPR'23) \cite{zhang2023domainadaptor} & 76.98$\pm$3.87 & 90.75$\pm$0.40 & 84.12$\pm$0.08 & 76.57$\pm$3.98 & 91.73$\pm$0.27 & 86.01$\pm$0.09 & 70.81$\pm$7.50 & 91.21$\pm$0.20 & 83.71$\pm$0.12 & 66.98$\pm$9.02 & 84.59$\pm$0.27 & 79.01$\pm$0.06 & 77.21$\pm$4.02 & 89.11$\pm$0.15 & 85.02$\pm$0.03 & 73.71 & 89.48 & 83.57 \\

        NC-TTT(CVPR'24) \cite{osowiechi2024nc} & 77.21$\pm$2.52 & 93.99$\pm$0.23 & 85.69$\pm$0.13 & 83.19$\pm$3.21 & \textbf{\textcolor{red}{93.57$\pm$0.21}} & \textbf{\textcolor{red}{86.94$\pm$0.07}} & 78.73$\pm$5.07 & 92.46$\pm$0.16 & 84.22$\pm$0.10 & 75.31$\pm$3.28 & 87.94$\pm$0.24 & 81.34$\pm$0.19 & 81.73$\pm$0.19 & 92.81$\pm$0.11 & 85.71$\pm$0.12 & 79.23 & 92.15 & 84.78 \\

        VPTTA(CVPR'24) \cite{chen2024each} & 75.57$\pm$4.14 & 92.64$\pm$0.04 & 85.71$\pm$0.01 & 79.42$\pm$3.87 & 92.12$\pm$0.05 & 85.14$\pm$0.02 & 71.69$\pm$2.48 & 92.97$\pm$0.06 & 83.02$\pm$0.02 & 64.48$\pm$5.61 & 86.14$\pm$0.11 & 77.82$\pm$0.09 & 75.83$\pm$2.59 & 90.81$\pm$0.10 & 85.15$\pm$0.08 & 73.40 & 91.34 & 83.77 \\

        GraTA(AAAI'25) \cite{chen2025gradient} & 79.39$\pm$2.23 & 92.07$\pm$0.11 & 84.79$\pm$0.20 & 82.51$\pm$2.23 & 92.19$\pm$0.17 & 85.39$\pm$0.07 & 78.94$\pm$2.23 & 92.05$\pm$0.27 & 84.13$\pm$0.12 & 76.75$\pm$2.23 & 90.79$\pm$0.34 & 83.17$\pm$0.08 & 75.71$\pm$2.23 & 89.02$\pm$0.10 & 82.29$\pm$0.07 & 78.66 & 91.22 & 83.95 \\

        TTDG(CVPR'25) \cite{lv2025test} & 83.49$\pm$2.23 & \textbf{\textcolor{red}{94.11$\pm$0.17}} & 87.41$\pm$0.17 & 83.13$\pm$3.02 & 92.67$\pm$0.27 & 86.11$\pm$0.19 & 83.68$\pm$1.88 & 93.51$\pm$0.41 & 87.21$\pm$0.14 & 79.34$\pm$4.31 & 91.01$\pm$0.31 & 84.30$\pm$0.08 & 84.78$\pm$3.01 & \textbf{\textcolor{red}{93.81$\pm$0.24}} & 87.47$\pm$0.10 & 82.88 & 93.02 & 86.50 \\

        \midrule

        SPEGC(Ours) & \textbf{\textcolor{red}{84.90$\pm$2.14}} & 93.81$\pm$0.37 & \textbf{\textcolor{red}{88.50$\pm$0.14}} & \textbf{\textcolor{red}{83.34$\pm$1.44}} & 93.01$\pm$0.24 & 86.74$\pm$0.27 & \textbf{\textcolor{red}{84.57$\pm$1.94}} & \textbf{\textcolor{red}{93.71$\pm$0.31}} & \textbf{\textcolor{red}{88.42$\pm$0.07}} & \textbf{\textcolor{red}{83.54$\pm$2.17}} & \textbf{\textcolor{red}{93.21$\pm$0.08}} & \textbf{\textcolor{red}{85.42$\pm$0.10}} & \textbf{\textcolor{red}{85.51$\pm$1.13}} & 93.54$\pm$0.18 & \textbf{\textcolor{red}{88.92$\pm$0.09}} & \textbf{\textcolor{red}{84.37}} & \textbf{\textcolor{red}{94.56}} & \textbf{\textcolor{red}{87.60}} \\
  
      \bottomrule 
    \end{tabular}
  
  } 
\end{table*}

\begin{table*}[t]
  \centering

  \captionsetup{justification=raggedright}
  \caption{
  Average performance (Mean ± Std.) of our SPEGC, "No adapt", and six SOTA methods on the polyp segmentation task, based on five experimental runs. "Domain A" denotes training on Domain A and testing on Domain B-D, and so forth for the other domains. The best results are highlighted in \textbf{\textcolor{red}{bold red}}.
  }
  \label{tab:Polyp}

  \resizebox{\textwidth}{!}{%

    \begin{tabular}{c| ccc| ccc| ccc| ccc| ccc}
      \toprule

      \multirow{2}{*}{Methods} & 

      \multicolumn{3}{c}{Domain A} & 
      \multicolumn{3}{c}{Domain B} & 
      \multicolumn{3}{c}{Domain C} & 
      \multicolumn{3}{c}{Domain D} & 

      \multicolumn{3}{c}{Average} \\

      \cmidrule(lr){2-4} \cmidrule(lr){5-7} \cmidrule(lr){8-10} \cmidrule(lr){11-13} \cmidrule(lr){14-16}

      & DSC & $E_{\text{max}}^{b}$ & $S_{\alpha}$ 
      & DSC & $E_{\text{max}}^{b}$ & $S_{\alpha}$ 
      & DSC & $E_{\text{max}}^{b}$ & $S_{\alpha}$ 
      & DSC & $E_{\text{max}}^{b}$ & $S_{\alpha}$ 
      & DSC $\uparrow$ & $E_{\text{max}}^{b} \uparrow$ & $S_{\alpha} \uparrow$ \\
      
      \midrule \midrule

        No Adapt (ResUNet-50) \cite{diakogiannis2020resunet} & 70.32 & 86.82 & 82.14 & 68.33 & 85.33 & 81.32 & 70.48 & 87.38 & 81.92 & 76.81 & 87.19 & 84.14 & 71.49 & 86.68 & 82.38 \\
        
        \midrule
        
        SAR(ICLR'23) \cite{niu2023towards} & 69.33$\pm$0.87 & 86.08$\pm$0.16 & 80.72$\pm$0.10 & 68.34$\pm$1.94 & 84.01$\pm$0.14 & 81.74$\pm$0.07 & 70.91$\pm$1.02 & 88.41$\pm$0.15 & 82.42$\pm$0.07 & 68.27$\pm$2.94 & 84.51$\pm$0.10 & 79.82$\pm$0.04 & 69.21 & 85.75 & 81.17 \\

        Domain Adaptor(CVPR'23) \cite{zhang2023domainadaptor} & 78.53$\pm$0.93 & 92.07$\pm$0.04 & 86.99$\pm$0.09 & 71.13$\pm$1.54 & 89.56$\pm$0.03 & 82.59$\pm$0.07 & 71.77$\pm$1.98 & 92.13$\pm$0.07 & 83.11$\pm$0.02 & 69.35$\pm$1.31 & 84.13$\pm$0.07 & 79.70$\pm$0.03 & 72.70 & 89.47 & 83.10 \\

        NC-TTT(CVPR'24) \cite{osowiechi2024nc} & 77.92$\pm$1.05 & 91.74$\pm$0.12 & 86.69$\pm$0.08 & \textbf{\textcolor{red}{73.77$\pm$2.06}} & \textbf{\textcolor{red}{91.72$\pm$0.02}} & \textbf{\textcolor{red}{83.48$\pm$0.06}} & 69.14$\pm$0.91 & 90.09$\pm$0.21 & 81.97$\pm$0.07 & 80.94$\pm$1.12 & 89.67$\pm$0.09 & 84.51$\pm$0.07 & 75.44 & 90.81 & 84.16 \\

        VPTTA(CVPR'24) \cite{chen2024each} & 77.68$\pm$0.51 & 92.61$\pm$0.05 & 87.01$\pm$0.07 & 71.76$\pm$0.62 & 88.82$\pm$0.04 & 82.14$\pm$0.09 & 71.68$\pm$0.58 & 91.98$\pm$0.04 & 82.17$\pm$0.07 & 80.77$\pm$0.24 & 90.17$\pm$0.09 & 84.71$\pm$0.10 & 73.40 & 90.90 & 84.01 \\

        GraTA(AAAI'25) \cite{chen2025gradient} & 79.33$\pm$2.13 & 93.17$\pm$0.14 & 88.59$\pm$0.19 & 70.23$\pm$2.04 & 88.56$\pm$0.13 & 82.09$\pm$0.17 & \textbf{\textcolor{red}{73.22$\pm$2.97}} & 92.75$\pm$0.16 & 83.51$\pm$0.18 & 82.19$\pm$1.82 & 89.91$\pm$0.14 & 85.67$\pm$0.18 & 76.24 & 91.01 & 84.97 \\

        TTDG(CVPR'25) \cite{lv2025test} & 82.90$\pm$1.13 & 94.00$\pm$0.09 & 89.82$\pm$0.12 & 70.57$\pm$1.21 & 89.01$\pm$0.07 & 81.51$\pm$0.09 & 70.30$\pm$1.88 & 91.83$\pm$0.11 & 83.32$\pm$0.06 & 81.05$\pm$1.31 & 89.33$\pm$0.07 & 85.30$\pm$0.10 & 76.20 & 91.04 & 84.99 \\

        \midrule

        SPEGC(Ours) & \textbf{\textcolor{red}{83.21$\pm$1.01}} & \textbf{\textcolor{red}{94.14$\pm$0.07}} & \textbf{\textcolor{red}{90.04$\pm$0.08}} & 72.87$\pm$0.90 & 89.20$\pm$0.11 & 82.42$\pm$0.06 & 72.85$\pm$0.87 & \textbf{\textcolor{red}{92.82$\pm$0.10}} & \textbf{\textcolor{red}{84.41$\pm$0.03}} & \textbf{\textcolor{red}{84.16$\pm$0.82}} & \textbf{\textcolor{red}{90.61$\pm$0.13}} & \textbf{\textcolor{red}{86.71$\pm$0.07}} & \textbf{\textcolor{red}{78.27}} & \textbf{\textcolor{red}{91.69}} & \textbf{\textcolor{red}{85.90}} \\
  
      \bottomrule 
    \end{tabular}
  
} 
\end{table*}

\noindent\textbf{Evaluation metrics.} We use three metrics to quantitatively evaluate segmentation performance. The main metric is the DICE Similarity Coefficient (DSC), which is used to quantify the degree of overlap between the predicted results and the labels. In addition, we also introduce two auxiliary metrics: the Enhanced Alignment metric ($E_\phi^{\max}$) \cite{fan2018enhanced} is used to evaluate the pixel-level and global similarity, and the Structural Similarity metric ($S_{\alpha}$) \cite{fan2017structure} is used to measure the structural consistency between predicted results and labels.

\begin{table*}[t]

  \centering 
  \captionsetup{justification=raggedright}
  \caption{Performance of our SPEGC and five competing methods on the OD/OC segmentation task under Long-term Continual Test-Time Adaptation. Numbers in \textbf{\textcolor{red}{bold red}} indicate the performance gain relative to the "No Adapt" baseline. Performance degradation is calculated as the difference between the overall average DSC and the first-round average DSC. "Ave": Abbreviation for "Average".}
  \label{tab:LCTTA} 
  \resizebox{\textwidth}{!}{%

    \begin{tabular}{c| cccccc| cccccc| cccccc| cccccc| cccccc| cc}
      \toprule 

      Time & \multicolumn{30}{c}{\tikz \draw[line width=1.5pt, -latex] (0,5ex) -- (36cm, 5ex);} & & \\

      \cmidrule(lr){1-33} 

      Round & \multicolumn{6}{c}{1} & \multicolumn{6}{c}{2} & \multicolumn{6}{c}{3} & \multicolumn{6}{c}{4} & \multicolumn{6}{c}{5} & \multicolumn{2}{c}{Average Performance} \\

      \cmidrule(lr){2-7} \cmidrule(lr){8-13} \cmidrule(lr){14-19} \cmidrule(lr){20-25} \cmidrule(lr){26-31} 

      Methods & A & B & C & D & E & Ave & A & B & C & D & E & Ave & A & B & C & D & E & Ave & A & B & C & D & E & Ave & A & B & C & D & E & Ave & DSC$\uparrow$ & Degra.$\downarrow$ \\
      \midrule \midrule

      No Adapt & 71.92 & 79.31 & 75.42 & 63.77 & 73.32 & 72.75 & 71.92 & 79.31 & 75.42 & 63.77 & 73.32 & 72.75 & 71.92 & 79.31 & 75.42 & 63.77 & 73.32 & 72.75 & 71.92 & 79.31 & 75.42 & 63.77 & 73.32 & 72.75 & 71.92 & 79.31 & 75.42 & 63.77 & 73.32 & 72.75 & 72.75 & -- \\
      
      \midrule

      SAR \cite{niu2023towards} & 74.23 & 80.23 & 72.22 & 70.19 & 70.35 & 73.44 & 73.97 & 80.12 & 71.73 & 69.61 & 69.87 & 73.06 & 73.49 & 79.47 & 71.42 & 69.01 & 69.17 & 72.51 & 73.22 & 78.94 & 70.88 & 68.49 & 68.61 & 72.03 & 72.01 & 78.28 & 70.11 & 68.03 & 68.14 & 71.31 & 72.47(\textbf{\textcolor{red}{-0.28}}) & 0.97 \\

      NC-TTT \cite{osowiechi2024nc}& 77.21 & 83.19 & 78.73 & 75.31 & 81.73 & 79.23 & 76.85 & 82.64 & 78.21 & 74.17 & 80.72 & 78.52 & 75.21 & 81.63 & 77.93 & 73.34 & 79.79 & 77.58 & 74.31 & 80.02 & 76.11 & 72.02 & 77.49 & 75.99 & 72.79 & 79.14 & 75.22 & 71.83 & 76.86 & 75.17 & 77.30(\textbf{\textcolor{red}{+4.55}}) & 1.93 \\

      VPTTA \cite{chen2024each}& 75.57 & 79.42 & 71.69 & 64.48 & 75.83 & 73.40 & 75.02 & 79.11 & 71.25 & 64.34 & 75.02 & 72.90 & 74.82 & 78.75 & 70.86 & 64.54 & 75.28 & 72.85 & 74.49 & 78.43 & 70.21 & 64.03 & 74.79 & 72.39 & 74.06 & 78.31 & 69.62 & 63.78 & 74.54 & 72.06 & 72.72(\textbf{\textcolor{red}{-0.03}}) & 0.68 \\

      GraTa \cite{chen2025gradient}& 79.39 & 82.51 & 78.94 & 76.75 & 75.71 & 78.66 & 77.47 & 81.23 & 77.02 & 74.93 & 76.28 & 77.39 & 75.14 & 78.91 & 74.86 & 72.58 & 74.19 & 75.14 & 73.02 & 76.46 & 72.73 & 71.41 & 72.92 & 73.31 & 71.11 & 74.38 & 70.19 & 69.82 & 70.44 & 71.19 & 75.13(\textbf{\textcolor{red}{+2.38}}) & 3.52 \\

      TTDG \cite{lv2025test}& 83.49 & 83.13 & 83.68 & 79.34 & 84.78 & 82.88 & 82.85 & 82.77 & 83.02 & 78.96 & 84.14 & 82.34 & 82.42 & 82.31 & 82.71 & 78.26 & 83.77 & 81.89 & 81.87 & 81.49 & 82.08 & 77.52 & 83.31 & 81.25 & 81.24 & 80.88 & 81.47 & 76.93 & 82.62 & 80.62 & 81.79(\textbf{\textcolor{red}{+9.04}}) & 1.08 \\
      
      \midrule
      Ours & 84.90 & 83.34 & 84.57 & 83.54 & 85.51 & 84.37 & 83.74 & 82.80 & 83.83 & 82.62 & 84.49 & 83.45 & 83.47 & 83.31 & 83.44 & 82.02 & 84.17 & 83.28 & 82.67 & 82.24 & 82.46 & 81.48 & 83.42 & 82.46 & 81.97 & 82.01 & 81.74 & 81.01 & 83.03 & 81.95 & 83.10(\textbf{\textcolor{red}{+10.35}}) & 1.27 \\
      \bottomrule 
    \end{tabular}%
    }
\end{table*}

\begin{table*}[t]
  \centering

  \captionsetup{justification=raggedright}
  \caption{Ablation study results on the OD/OC segmentation task. The best in each column is marked in \textbf{bold}.}
  \label{tab:ablation} 

  \resizebox{0.80\textwidth}{!}{%

    \begin{tabular}{cccc|ccccc|c}
      \toprule

      \multicolumn{4}{c|}{Methods} & Domain A & Domain B & Domain C & Domain D & Domain E & Average \\
      
      \midrule

      Clustering & MC Dropout & CO-Prompt+$L_{C}$ & HE-Prompt & DSC $\uparrow$ & DSC $\uparrow$ & DSC $\uparrow$ & DSC $\uparrow$ & DSC $\uparrow$ & DSC $\uparrow$ \\
      
      \midrule \midrule

      & & & & 71.92 & 79.31 & 75.42 & 63.77 & 73.32 & 72.75 \\
      
      \checkmark & & & & 73.49 & 80.94 & 78.09 & 64.47 & 76.19 & 74.64 \\
      
      \checkmark & \checkmark & & & 74.02 & 82.27 & 77.58 & 68.91 & 79.82 & 76.52 \\
      
      \checkmark & \checkmark & & \checkmark & 74.14 & 80.21 & 77.22 & 67.31 & 78.08 & 75.39 \\
      
      \checkmark & \checkmark & \checkmark & & 77.09 & \textbf{84.24} & 82.50 & 78.32 & 83.18 & 81.07 \\
      
      \checkmark & \checkmark & \checkmark& \checkmark & \textbf{84.90} & 83.34 & \textbf{84.57} & \textbf{83.54} & \textbf{85.51} & \textbf{84.37} \\
  
      \bottomrule 
    \end{tabular}
    
  } 
\end{table*}

\vspace{-1mm}
\subsection{Comparative Experiments}
\vspace{-1mm}

All experiments employ ResUNet-50 \cite{diakogiannis2020resunet} as the common segmentation architecture. We establish a "No Adapt" baseline, which evaluates the source-trained model directly on target domains without adaptation, to quantify the initial domain gap. SPEGC is benchmarked against six representative state-of-the-art (SOTA) adaptation techniques, which we categorize in terms of their core mechanisms: (1) entropy-based: SAR \cite{niu2023towards}; (2) batch normalization-based: DomainAdaptor \cite{zhang2023domainadaptor} and VPTTA \cite{chen2024each}; (3) noise estimation: NC-TTT \cite{osowiechi2024nc}; (4) gradient alignment: GraTa \cite{chen2025gradient}; and (5) graph matching: TTDG \cite{lv2025test}. We follow a rigorous cross-domain evaluation protocol: each domain is sequentially used to train the source model, while all remaining domains are treated as unseen, continual target streams for adaptation and evaluation.

\noindent\textbf{Comparison on the OD/OC segmentation task.} \cref{tab:ODOC} presents the quantitative comparison results of our proposed SPEGC, the "No Adapt" baseline, and six competing TTA methods on the OD/OC segmentation task. As shown in \cref{tab:ODOC}, all TTA methods outperform the "No Adapt" baseline, validating the necessity of adaptation to mitigate distribution shifts. Our proposed SPEGC achieves the best average performance across all target domains. Specifically, our method achieves an average DSC improvement of 1.49\% over the next-best method, TTDG \cite{lv2025test}, and shows consistent gains across all evaluated domains.

\noindent\textbf{Comparison on the polyp segmentation task.} Further experiments were conducted on the polyp segmentation task (\cref{tab:Polyp}). Notably, unlike the retinal task, methods that rely on entropy minimization (e.g., SAR \cite{niu2023towards}) exhibit significant performance degradation, even falling below the "No Adapt" baseline. We speculate this decay is attributable to the subtle, "cryptic" nature of polyp targets. Under the influence of distribution shifts, entropy minimization methods tend to converge to over-confident, erroneous predictions, thereby generating misleading gradients during the adaptation process. In contrast, our proposed SPEGC relies on an optimization approach that leverages the internal structure of the data rather than the own prediction confidence of the model, effectively circumventing the pitfalls of entropy minimization. Consequently, SPEGC surpasses all competing methods in this more challenging task.

\noindent\textbf{Evaluation under long-term continual test-time adaptation.} Beyond single-round adaptation, model stability and robustness under the Long-term Continual Test-Time Adaptation (L-CTTA) \cite{wang2022continual} setting are crucial. We conducted five consecutive rounds of experiments on the optic disc/cup segmentation task, requiring the model to learn continuously across domains without parameter resets. DomainAdaptor \cite{zhang2023domainadaptor} is excluded as it resets the model before each step. We compare our SPEGC against SAR \cite{niu2023towards} , NC-TTT \cite{osowiechi2024nc} , VPTTA \cite{chen2024each} , GraTa \cite{chen2025gradient}, and TTDG \cite{lv2025test}, evaluating long-term stability using two metrics : (1) performance degradation in the source domain (evaluating catastrophic forgetting) ; and (2) overall average performance across all rounds (evaluating error accumulation). Results are presented in \cref{tab:LCTTA}.

As shown in \cref{tab:LCTTA}, strategies employing model resetting (e.g., SAR \cite{niu2023towards} ) or priors (e.g., TTDG \cite{lv2025test} ) effectively mitigate error, exhibiting low performance degradations of 0.97\% and 1.08\%, respectively. VPTTA \cite{chen2024each} , which freezes source weights, shows minimal degradation (0.68\%) as its parameters are invariant. In contrast, GraTa \cite{chen2025gradient}, which is based on entropy minimization, suffers a significant performance decay (3.52\%). Throughout the evaluation, SPEGC demonstrates robust adaptability, achieving the SOTA overall average performance (DSC 83.10\% ), proving its superior ability to mitigate error accumulation. Furthermore, it effectively limits catastrophic forgetting, incurring a performance degradation of only 1.27\%. More visualization results are available in the supplementary materials.

\vspace{-1mm}
\subsection{Component Effectiveness Analysis}

\textbf{Ablation study.} To validate the effectiveness and contribution of the individual key components within our proposed SPEGC framework, we conduct a series of ablation experiments on the OD/OC segmentation task. The results are presented in \cref{tab:ablation}. The experiment specifically investigates the following three aspects. (1) Differentiable Graph Clustering: To evaluate the baseline performance achieved by only using the refined edge similarity matrix. (2) Uncertainty Sampling: To validate using MC Dropout \cite{gal2016dropout} to select low-uncertainty nodes for high-quality graph construction. (3) Semantic Prompting: To evaluate the individual contributions of the commonality and heterogeneity prompts, as well as their synergistic effect.

In \cref{tab:ablation}, the "No Adapt" baseline achieves an average DSC of only 72.75\%. By introducing only the differentiable graph clustering, performance increases to 74.64\%. Building upon this, introducing MC Dropout \cite{gal2016dropout} for uncertainty sampling yields a significant boost to 76.52\%. This validates the critical importance of sampling reliable, low-uncertainty nodes to construct a high-quality graph structure. Starting from the 76.52\% baseline, we investigate the prompting modules. Adding only the heterogeneity prompts (without loss constraints) conversely drops performance to 75.39\%, indicating that unconstrained prompts introduce noise. In contrast, introducing commonality prompts, enhanced by the clustering loss, substantially increases performance to 81.07\%. This highlights the value of injecting constrained, cross-domain shared context for adaptation. Finally, our complete model achieves the best performance of 84.37\%, significantly surpassing any single-prompt variant. This demonstrates the value of the non-shared, domain-specific information as a beneficial complement.

\begin{figure}[h]
  \centering
  \includegraphics[width=1\linewidth]{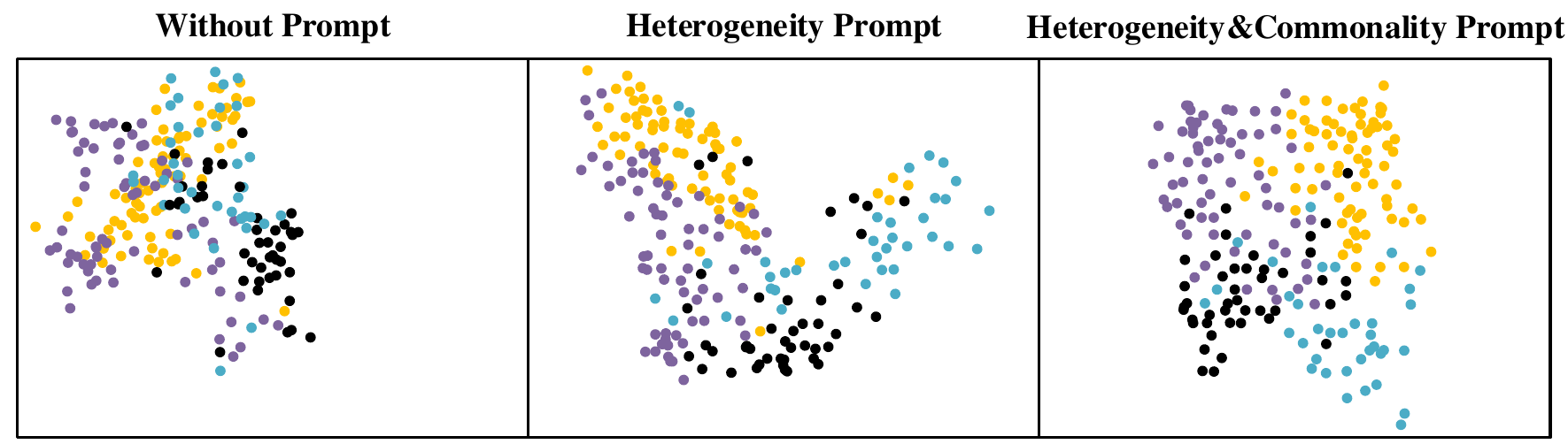}
  
  \caption{
  t-SNE \cite{maaten2008visualizing} visualization of the embeddings learned by D-Prompt and SEC-Prompt on the OD/OC task. Different colors represent images from different domains.
  }
  
  \label{fig:TSen}
\end{figure}

\noindent\textbf{Prompt analysis.} To validate our prompt design, we use t-SNE \cite{maaten2008visualizing} to visualize the features of the OD/OC task (source A, target B-E). In \cref{fig:TSen}, where colors represent domains, features without prompt enhancement are tightly intermingled with indistinct boundaries. Using only the heterogeneity prompts causes the features to disperse and form clearer boundaries. After incorporating the commonality prompts, a significant reduction in intra-class distance is observed, and the features become more compactly clustered.

\vspace{-1mm}

\vspace{-6mm}

\subsection{Hyperparameter Analysis}

To investigate the sensitivity of the model's performance to key hyperparameters, this section conducts a series of analytical experiments on the OD/OC segmentation task, focusing on the feature pool size, the number of clusters (Z), and the number of prompts (M).

\begin{figure}[h]
  \centering
  \includegraphics[width=0.97\linewidth]{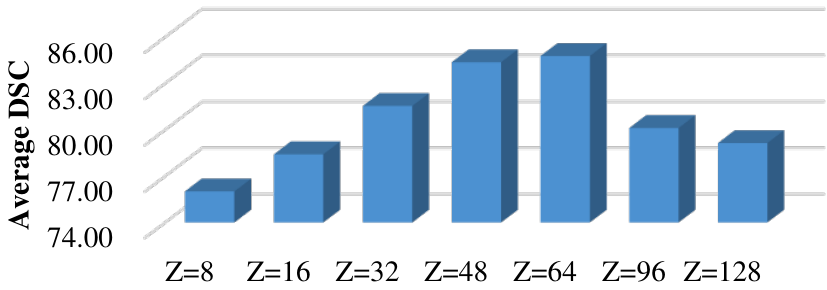}
  
  \caption{
  Performance of SPEGC with various Z on the OD/OC segmentation task.
  }

  \label{fig:Z}
\end{figure}

\begin{figure}[h]
  \centering

  \includegraphics[width=0.97\linewidth]{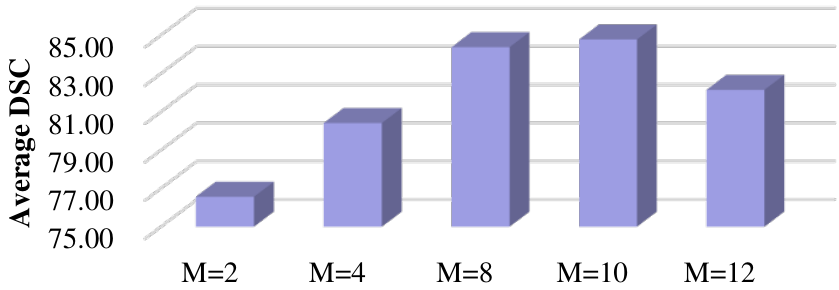}
  
  \caption{
  Performance of our SPEGC with various M on the OD/OC segmentation task.
  }
  
  \label{fig:M}
\end{figure}

\noindent\textbf{Clusters ($Z$) and prompt pool ($M$) analysis.} \cref{fig:Z} and \cref{fig:M} illustrate the impact of $Z$ and $M$, respectively. When analyzing $Z$ (fixing $M=8$), \cref{fig:Z} shows performance peaks at $Z=64$ and is slightly lower at $Z=48$. $Z$ that is too small may erroneously merge distinct categories, while one that is too large may fragment coherent semantic features. When analyzing $M$ (fixing $Z=48$), \cref{fig:M} shows comparable peak performance at $M=8$ and $M=10$. Balancing the trade-off between efficiency and performance, we select $Z=48$ and $M=8$ as the final configuration.

\vspace{-5mm}

\vspace{-3mm}

\noindent\textbf{Feature pool analysis.} Fixing $Z=48$ and $M=8$, the results are shown in \cref{tab:feature_pool_comparison}. The results show SPEGC achieves peak performance (DSC 85.24\%) with a feature pool size of 7, nearly 1\% higher than size 3 (DSC 84.37\%). However, since the affinity matrix computation scales quadratically with the number of nodes, increasing the pool size from 3 to 7 results in a nearly 4-fold increase in FLOPS (from 5.8G to 21.7G). A pool that is too small fails to capture sufficient semantics, while one that is too large introduces redundancy. Balancing the trade-off between performance and efficiency, we select 3 as the best configuration of the feature pool.

\vspace{-2mm}

\begin{table}[h] 
    \centering 
    \captionsetup{justification=raggedright}
    \caption{
    Ablation study of feature pool size on the OD/OC task. "Avg. DSC" denotes the average DSC across the five sites, "time" indicates the inference time per image, and "VRAM" represents the GPU memory usage.
    }
    \label{tab:feature_pool_comparison}
    
    \resizebox{\linewidth}{!}{
    \begin{tabular}{c|c|c|c|c} 
    \toprule 

    Feature Pool & Avg. DSC $\uparrow$ & FLOPs(G) $\downarrow$ & Time(s/img) $\downarrow$ & VRAM(MB) $\downarrow$ \\
    \midrule \midrule

    1 & 79.21 & 2.917 & 0.314 & 2798 \\
    \midrule
    3 & 84.37 & 5.817 & 0.521 & 3276 \\
    \midrule
    7 & 85.24 & 21.72 & 1.471 & 4817 \\
    \midrule
    15 & 80.14 & 120.3 & 4.774 & 8738 \\
    
    \bottomrule
    \end{tabular}
    } 
\end{table}

\vspace{-3mm}

%You must include your signed IEEE copyright release form when you submit your finished paper.
%We MUST have this form before your paper can be published in the proceedings.

%Please direct any questions to the production editor in charge of these proceedings at the IEEE Computer Society Press:
%\url{https://www.computer.org/about/contact}.

%% file: sec/5_conclusion.tex
\vspace{-1mm}
\section{Conclusion}
In this work, we have addressed the critical challenge of CTTA for medical image segmentation. We proposed SPEGC, a novel framework that moves beyond error-prone entropy minimization by leveraging high-order structural abstractions from the data itself. Extensive experiments on two challenging medical segmentation benchmarks demonstrate that SPEGC significantly outperforms SOTA methods. Crucially, L-CTTA evaluation confirms its superior ability to mitigate both catastrophic forgetting and the accumulation of errors, validating the robustness of our structural adaptation. 

We acknowledge that explicit graph construction in DGCS, while effective, introduces a higher computational cost than lighter-weight adaptation methods. Our future work will focus on optimizing this component, perhaps by exploring graph sparsification techniques or more efficient optimal transport solvers.

\section*{Acknowledgements}
This work was supported in part by the National Natural Science Foundation of China (Grant Nos. 62271296 and 62201334), the Young Science and Technology Innovation Leading Talents Program of Xi'an City (Grant No. 25ZQRC00019), the Innovation Capability Support Plan Project in Shaanxi Province (Grant No. 2025RS-CXTD-012), the Scientific Research Program Funded by Shaanxi Provincial Education Department (Grant Nos. 23JP022, 23JP014, 24JK0350, and 25JP023), and the Natural Science Basic Research Program of Shaanxi (Grant No. 2025JC-YBQN-800).

%% file: sec/A1_suppl.tex
\clearpage
\setcounter{page}{1}
\maketitlesupplementary

\appendix

\setcounter{figure}{0}                       
\setcounter{table}{0}                        
\renewcommand{\thefigure}{A.\arabic{figure}} 
\renewcommand{\thetable}{A.\arabic{table}}   

\section{Theoretical Analysis}
\label{sec:Theoretical Analysis}

\begin{algorithm*}[t]
\caption{SPEGC: Continual Test-Time Adaptation via Semantic-Prompt-Enhanced Graph Clustering}
\label{alg:spegc}
\begin{algorithmic}[1]
\State \textbf{Initialize:} Source-trained model $f_{\sigma}$ with parameters $\sigma \leftarrow \sigma_S$;
\Statex Learnable prompt pools $P_{CO} \in \mathbb{R}^{M \times h}$, $P_{HE} \in \mathbb{R}^{M \times h}$;
\Statex Learnable projections $W_q \in \mathbb{R}^{h \times h}, W_k \in \mathbb{R}^{h \times h}$;
\Statex Learnable context vector $c_p \in \mathbb{R}^{h}$;
\Statex Feature queue $\mathcal{Q} \leftarrow \emptyset$ (max capacity $N$);

\State \textbf{Input:} Continuous target domain image stream $\{x_i\}_{i=1}^{\infty}$.
\State \textbf{Output:} Adapted prediction stream $\{O_i\}_{i=1}^{\infty}$.

\For{$i \in [1,m]$}
    \Statex \textbf{1. Semantic Prompt Feature Enhancement (SPFE)}
    \State Generate $t$ stochastic feature maps $\{F_k\}_{k=1}^t \leftarrow f_{\sigma}(x_i)$ using MC Dropout.
    \State Estimate uncertainty $U \leftarrow \operatorname{Variance}(\{F_k\}_{k=1}^t)$ (\cref{eq:1}).
    \State Select $n_i$ low-uncertainty nodes $\mathcal{V}_i \leftarrow \operatorname{SampleLowUncertainty}(U, \{F_k\}, p\%)$.
    \State Aggregate node features into global query $\hat{q}_i \leftarrow \operatorname{AttentionPooling}(\mathcal{V}_i, c_p)$ (\cref{eq:2}).
    \State Retrieve commonality prompt $p_{CO}(i) \leftarrow \operatorname{ReverseAttention}(\hat{q}_i, P_{CO})$ (\cref{eq:3,eq:5}).
    \State Retrieve heterogeneity prompt $p_{HE}(i) \leftarrow \operatorname{Attention}(\hat{q}_i, P_{HE})$ (\cref{eq:4,eq:6}).
    \State Obtain enhanced features $\mathcal{V}_i^* \leftarrow \mathcal{V}_i + p_{CO}(i) + p_{HE}(i)$ (\cref{eq:7}).
    
    \Statex \textbf{2. Differentiable Graph Clustering Solver (DGCS)}
    \State Assemble pseudo-batch $\mathcal{V}^* \leftarrow \mathcal{Q} \cup \{\mathcal{V}_i^*\}$.
    \State Update feature queue: $\mathcal{Q}.\text{enqueue}(\mathcal{V}_i^*)$.
    \If{$|\mathcal{Q}| > N$}
        \State $\mathcal{Q}.\text{dequeue}()$.
    \EndIf
    
    \State Calculate global similarity $S \in \mathbb{R}^{V \times V}$ (\cref{eq:eq8}), where $V$ is total nodes in $\mathcal{V}^*$.
    \State Determine node densities $D(v_i) \leftarrow \sum_{j} \operatorname{ReLU}(S(i, j))$ (\cref{eq:9}).
    \State Define density-aware edge similarity $S'(i,j)$ (\cref{eq:10}).
    \State Formulate optimal transport cost matrix $\mathbf{D}$ from $S'$ (costs for selecting/rejecting $k=V-Z$ edges) (\cref{eq:11,eq:12}).
    \State Solve for optimal transport plan $\Gamma^* \leftarrow \operatorname{Sinkhorn}(\mathbf{D}, \theta)$ (\cref{eq:13,eq:14}).
    \State Extract refined edge similarity $S^* \leftarrow \operatorname{Reshape}(\Gamma_{:,2}^*)$.
    
    \Statex \textbf{3. Joint Optimization \& Adaptation}
    \State Acquire semantic predictions $P \in \mathbb{R}^{V \times C}$ for nodes in $\mathcal{V}^*$ from $f_{\sigma}$.
    \State Calculate graph consistency loss $L_G$ (\cref{eq:16}) and clustering loss $L_C$ (\cref{eq:17}).
    \State Determine total loss $L \leftarrow L_G + \lambda L_C$ (\cref{eq:15}).
    \State Update all learnable parameters $\{\sigma, P_{CO}, P_{HE}, W_q, W_k, c_p\}$ by backpropagating $L$.
    
    \Statex \textbf{4. Inference}
    \State Generate final prediction $O_i \leftarrow f_{\sigma}(x_i)$ (using the updated parameters $\sigma$).
\EndFor{\textbf{end for}}
\end{algorithmic}
\end{algorithm*}

\begin{table*}[t]
  \centering

  \captionsetup{justification=raggedright}
  \caption{
Quantitative comparison under \textbf{Mixed Distribution Shifts} on the OD/OC segmentation task. Models are trained on the indicated source domain and adapted to a composite target stream formed by shuffling samples from all remaining domains. We report the Mean $\pm$ Std. over five independent runs. 
The best results are highlighted in \textbf{\textcolor{red}{bold red}}.
  }
  \label{tab:mix_ODOC}

  \resizebox{\textwidth}{!}{%

    \begin{tabular}{c| ccc| ccc| ccc| ccc| ccc| ccc}
      \toprule

      \multirow{2}{*}{Methods} & 

      \multicolumn{3}{c}{Domain A} & 
      \multicolumn{3}{c}{Domain B} & 
      \multicolumn{3}{c}{Domain C} & 
      \multicolumn{3}{c}{Domain D} & 
      \multicolumn{3}{c}{Domain E} & 

      \multicolumn{3}{c}{Average} \\

      \cmidrule(lr){2-4} \cmidrule(lr){5-7} \cmidrule(lr){8-10} \cmidrule(lr){11-13} \cmidrule(lr){14-16} \cmidrule(lr){17-19}

      & DSC & $E_\phi^{\max}$ & $S_{\alpha}$ 
      & DSC & $E_\phi^{\max}$ & $S_{\alpha}$ 
      & DSC & $E_\phi^{\max}$ & $S_{\alpha}$ 
      & DSC & $E_\phi^{\max}$ & $S_{\alpha}$ 
      & DSC & $E_\phi^{\max}$ & $S_{\alpha}$ 
      & DSC $\uparrow$ & $E_{\text{max}}^{b} \uparrow$ & $S_{\alpha} \uparrow$ \\
      
      \midrule \midrule

        No Adapt (ResUNet-50) \cite{diakogiannis2020resunet} & 71.92 & 88.67 & 80.84 & 79.31 & 90.42 & 84.82 & 75.42 & 89.24 & 81.62 & 63.77 & 85.49 & 78.28 & 73.32 & 89.67 & 81.81 & 72.75 & 88.70 & 81.47 \\
        
        \midrule
        
        SAR(ICLR'23) \cite{niu2023towards} & 74.03$\pm$6.43 & 90.07$\pm$0.31 & 83.52$\pm$0.17 & 80.33$\pm$2.42 & 92.86$\pm$0.31 & 85.29$\pm$0.19 & 71.42$\pm$4.67 & 91.55$\pm$0.12 & 83.01$\pm$0.16 & 69.89$\pm$1.32 & 86.01$\pm$0.14 & 79.17$\pm$0.98 & 69.75$\pm$5.61 & 88.03$\pm$0.97 & 86.12$\pm$0.47 & 73.08 & 89.70 & 83.42 \\

        Domain Adaptor(CVPR'23) \cite{zhang2023domainadaptor} & 76.28$\pm$4.47 & 90.75$\pm$0.40 & 83.82$\pm$0.14 & 76.27$\pm$4.28 & 91.03$\pm$0.37 & 85.71$\pm$0.14 & 70.21$\pm$8.12 & 91.01$\pm$0.31 & 82.44$\pm$0.24 & 66.18$\pm$9.82 & 83.41$\pm$0.42 & 78.34$\pm$0.17 & 76.31$\pm$4.42 & 88.71$\pm$0.21 & 84.62$\pm$0.08 & 73.05 & 88.98 & 82.99 \\

        NC-TTT(CVPR'24) \cite{osowiechi2024nc} & 76.81$\pm$2.12 & 92.79$\pm$0.23 & 85.69$\pm$0.13 & 82.74$\pm$3.48 & \textbf{\textcolor{red}{93.47$\pm$0.34}} & \textbf{\textcolor{red}{86.71$\pm$0.10}} & 77.93$\pm$6.07 & 92.26$\pm$0.21 & 84.02$\pm$0.14 & 75.05$\pm$4.12 & 87.74$\pm$0.31 & 81.54$\pm$0.39 & 81.23$\pm$0.39 & 92.61$\pm$0.14 & 85.01$\pm$0.12 & 78.75 & 91.77 & 84.59 \\

        VPTTA(CVPR'24) \cite{chen2024each} & 75.17$\pm$5.01 & 92.44$\pm$0.10 & 85.71$\pm$0.01 & 79.02$\pm$3.94 & 92.01$\pm$0.07 & 84.84$\pm$0.06 & 71.41$\pm$2.56 & 92.24$\pm$0.10 & 82.79$\pm$0.06 & 64.02$\pm$6.02 & 85.89$\pm$0.14 & 77.61$\pm$0.11 & 75.73$\pm$3.41 & 90.72$\pm$0.14 & 84.41$\pm$0.11 & 73.07 & 90.66 & 83.08 \\

        GraTA(AAAI'25) \cite{chen2025gradient} & 78.14$\pm$4.49 & 91.98$\pm$0.08 & 83.51$\pm$0.10 & 81.21$\pm$3.03 & 91.71$\pm$0.27 & 84.14$\pm$0.14 & 77.02$\pm$3.43 & 91.57$\pm$0.41 & 83.89$\pm$0.42 & 74.15$\pm$3.78 & 90.15$\pm$0.34 & 82.46$\pm$0.17 & 74.79$\pm$3.57 & 88.62$\pm$0.18 & 82.00$\pm$0.13 & 77.06 & 90.08 & 83.20 \\

        TTDG(CVPR'25) \cite{lv2025test} & 82.74$\pm$3.10 & \textbf{\textcolor{red}{94.02$\pm$0.14}} & 86.81$\pm$0.19 & \textbf{\textcolor{red}{82.91$\pm$3.42}} & 92.09$\pm$0.21 & 85.99$\pm$0.12 & 82.97$\pm$2.14 & \textbf{\textcolor{red}{93.47$\pm$0.32}} & 87.00$\pm$0.18 & 78.74$\pm$4.07 & 91.14$\pm$0.28 & 83.97$\pm$0.10 & 84.51$\pm$3.27 & \textbf{\textcolor{red}{93.64$\pm$0.27}} & 87.12$\pm$0.14 & 82.37 & 92.82 & 86.18 \\

        \midrule

        SPEGC(Ours) & \textbf{\textcolor{red}{84.72$\pm$2.30}} & 93.64$\pm$0.47 & \textbf{\textcolor{red}{88.09$\pm$0.24}} & 82.88$\pm$1.79 & 92.89$\pm$0.26 & 86.41$\pm$0.37 & \textbf{\textcolor{red}{83.79$\pm$2.41}} & 93.42$\pm$0.40 & \textbf{\textcolor{red}{88.02$\pm$0.17}} & \textbf{\textcolor{red}{83.04$\pm$2.57}} & \textbf{\textcolor{red}{93.04$\pm$0.11}} & \textbf{\textcolor{red}{85.32$\pm$0.14}} & \textbf{\textcolor{red}{85.02$\pm$2.07}} & 93.34$\pm$0.28 & \textbf{\textcolor{red}{88.81$\pm$0.13}} & \textbf{\textcolor{red}{83.89}} & \textbf{\textcolor{red}{93.47}} & \textbf{\textcolor{red}{87.33}} \\
  
      \bottomrule 
    \end{tabular}
  
  } 
\end{table*}

\begin{table*}[t]
  \centering

  \captionsetup{justification=raggedright}
  \caption{
Quantitative comparison under \textbf{Mixed Distribution Shifts} on the polyp segmentation task. Models are trained on the indicated source domain and adapted to a composite target stream formed by shuffling samples from all remaining domains. We report the Mean $\pm$ Std. over five independent runs. The best results are highlighted in \textbf{\textcolor{red}{bold red}}.
  }
  \label{tab:mix_Polyp}

  \resizebox{\textwidth}{!}{%

    \begin{tabular}{c| ccc| ccc| ccc| ccc| ccc}
      \toprule 

      \multirow{2}{*}{Methods} & 

      \multicolumn{3}{c}{Domain A} & 
      \multicolumn{3}{c}{Domain B} & 
      \multicolumn{3}{c}{Domain C} & 
      \multicolumn{3}{c}{Domain D} & 

      \multicolumn{3}{c}{Average} \\

      \cmidrule(lr){2-4} \cmidrule(lr){5-7} \cmidrule(lr){8-10} \cmidrule(lr){11-13} \cmidrule(lr){14-16}

      & DSC & $E_{\text{max}}^{b}$ & $S_{\alpha}$ 
      & DSC & $E_{\text{max}}^{b}$ & $S_{\alpha}$ 
      & DSC & $E_{\text{max}}^{b}$ & $S_{\alpha}$ 
      & DSC & $E_{\text{max}}^{b}$ & $S_{\alpha}$ 
      & DSC $\uparrow$ & $E_{\text{max}}^{b} \uparrow$ & $S_{\alpha} \uparrow$ \\
      
      \midrule \midrule

        No Adapt (ResUNet-50) \cite{diakogiannis2020resunet} & 70.32 & 86.82 & 82.14 & 68.33 & 85.33 & 81.32 & 70.48 & 87.38 & 81.92 & 76.81 & 87.19 & 84.14 & 71.49 & 86.68 & 82.38 \\
        
        \midrule
        
        SAR(ICLR'23) \cite{niu2023towards} & 69.01$\pm$0.98 & 85.79$\pm$0.14 & 80.21$\pm$0.14 & 67.81$\pm$2.32 & 83.87$\pm$0.07 & 81.51$\pm$0.14 & 70.51$\pm$1.47 & 88.31$\pm$0.17 & 82.40$\pm$0.09 & 68.02$\pm$2.44 & 84.37$\pm$0.14 & 79.68$\pm$0.06 & 68.84 & 85.59 & 80.95 \\

        Domain Adaptor(CVPR'23) \cite{zhang2023domainadaptor} & 78.17$\pm$1.14 & 91.84$\pm$0.07 & 86.41$\pm$0.12 & 70.87$\pm$1.74 & 89.36$\pm$0.05 & 82.13$\pm$0.12 & 71.63$\pm$2.04 & 91.98$\pm$0.12 & 83.04$\pm$0.04 & 68.76$\pm$1.14 & 83.97$\pm$0.07 & 79.52$\pm$0.04 & 72.36 & 89.29 & 82.78 \\

        NC-TTT(CVPR'24) \cite{osowiechi2024nc} & 77.64$\pm$1.31 & 91.54$\pm$0.17 & 86.52$\pm$0.12 & \textbf{\textcolor{red}{73.51$\pm$2.34}} & \textbf{\textcolor{red}{91.64$\pm$0.04}} & \textbf{\textcolor{red}{83.17$\pm$0.04}} & 68.93$\pm$1.02 & 89.81$\pm$0.17 & 81.63$\pm$0.12 & 80.67$\pm$1.04 & 89.51$\pm$0.14 & 84.28$\pm$0.13 & 75.19 & 90.63 & 83.90 \\

        VPTTA(CVPR'24) \cite{chen2024each} & 77.38$\pm$0.68 & 92.11$\pm$0.09 & 86.76$\pm$0.10 & 71.48$\pm$0.87 & 88.42$\pm$0.07 & 81.97$\pm$0.13 & 71.01$\pm$0.82 & 91.64$\pm$0.08 & 81.93$\pm$0.10 & 80.42$\pm$0.29 & 89.93$\pm$0.11 & 84.54$\pm$0.07 & 75.07 & 90.53 & 83.80 \\

        GraTA(AAAI'25) \cite{chen2025gradient} & 78.67$\pm$3.78 & 92.54$\pm$0.19 & 87.81$\pm$0.21 & 68.82$\pm$2.62 & 87.97$\pm$0.19 & 81.54$\pm$0.12 & \textbf{\textcolor{red}{72.81$\pm$3.77}} & 92.22$\pm$0.21 & 82.82$\pm$0.21 & 81.52$\pm$2.96 & 89.33$\pm$0.21 & 85.19$\pm$0.14 & 75.45 & 90.51 & 84.54 \\

        TTDG(CVPR'25) \cite{lv2025test} & 82.21$\pm$1.46 & 93.69$\pm$0.11 & \textbf{\textcolor{red}{89.77$\pm$0.19}} & 70.02$\pm$1.64 & 88.72$\pm$0.12 & 81.13$\pm$0.06 & 69.82$\pm$2.37 & 91.63$\pm$0.14 & 83.08$\pm$0.08 & 80.54$\pm$1.27 & 89.14$\pm$0.05 & 85.16$\pm$0.11 & 75.65 & 90.78 & 84.74 \\

        \midrule

        SPEGC(Ours) & \textbf{\textcolor{red}{83.04$\pm$1.14}} & \textbf{\textcolor{red}{94.03$\pm$0.08}} & 89.72$\pm$0.09 & 72.62$\pm$0.86 & 89.06$\pm$0.14 & 82.17$\pm$0.08 & 72.51$\pm$0.63 & \textbf{\textcolor{red}{92.61$\pm$0.08}} & \textbf{\textcolor{red}{84.32$\pm$0.04}} & \textbf{\textcolor{red}{83.82$\pm$0.93}} & \textbf{\textcolor{red}{90.48$\pm$0.11}} & \textbf{\textcolor{red}{86.32$\pm$0.09}} & \textbf{\textcolor{red}{78.00}} & \textbf{\textcolor{red}{91.55}} & \textbf{\textcolor{red}{85.68}} \\
  
      \bottomrule 
    \end{tabular}
  
} 
\end{table*}

\begin{figure*}[h]
  \centering
  \includegraphics[width=1\linewidth]{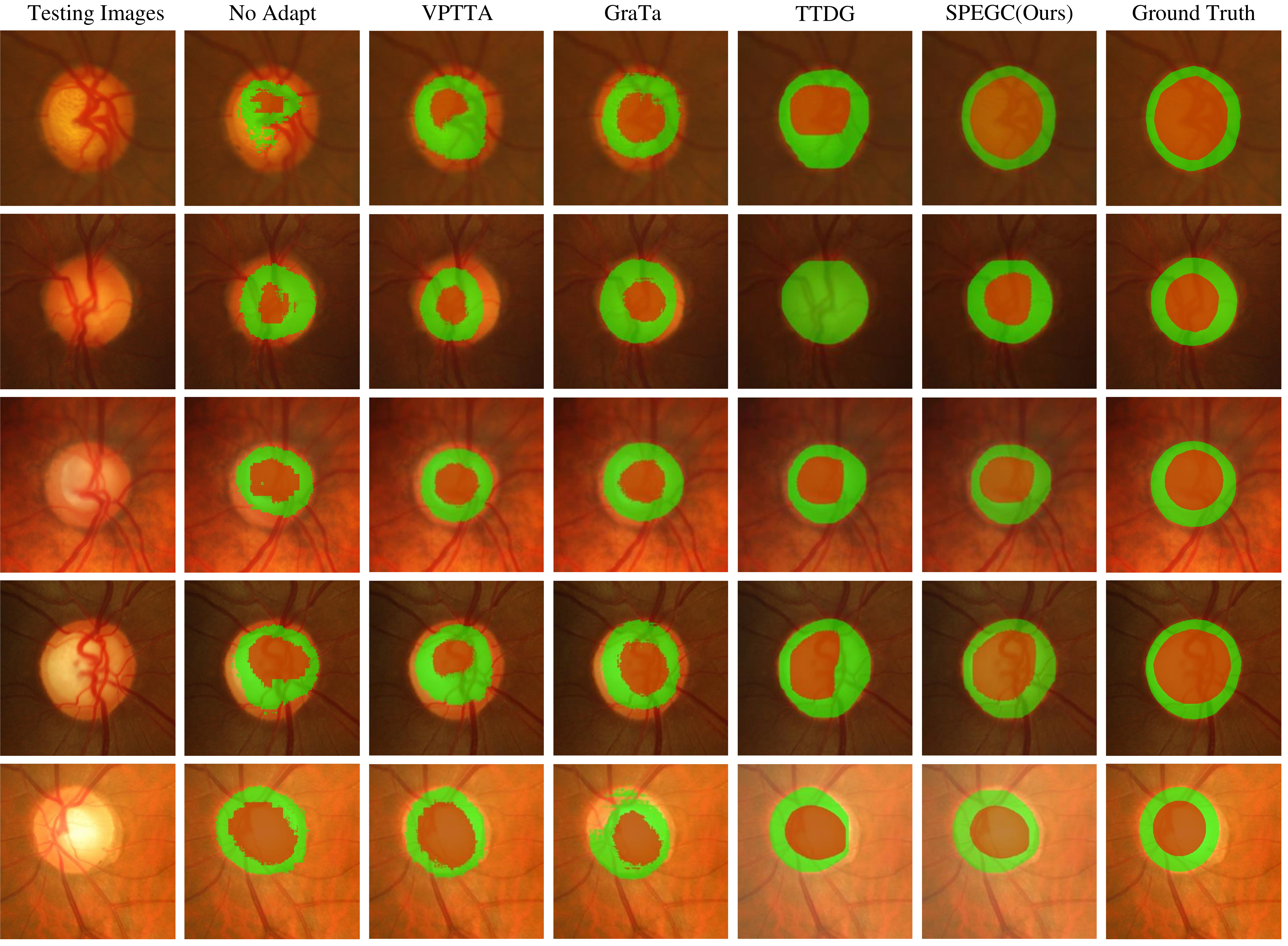}

  \caption{
Visualization comparison of segmentation results for the No Adapt baseline, VPTTA \cite{chen2024each}, GraTa \cite{chen2025gradient}, TTDG \cite{lv2025test}, and SPEGC(Ours) in retinal fundus segmentation. Different colors represent the segmentation instances of different classes identified by the network.
  }
  \label{fig:V_F}
\end{figure*}

\begin{figure*}[h]
  \centering
  \includegraphics[width=1\linewidth]{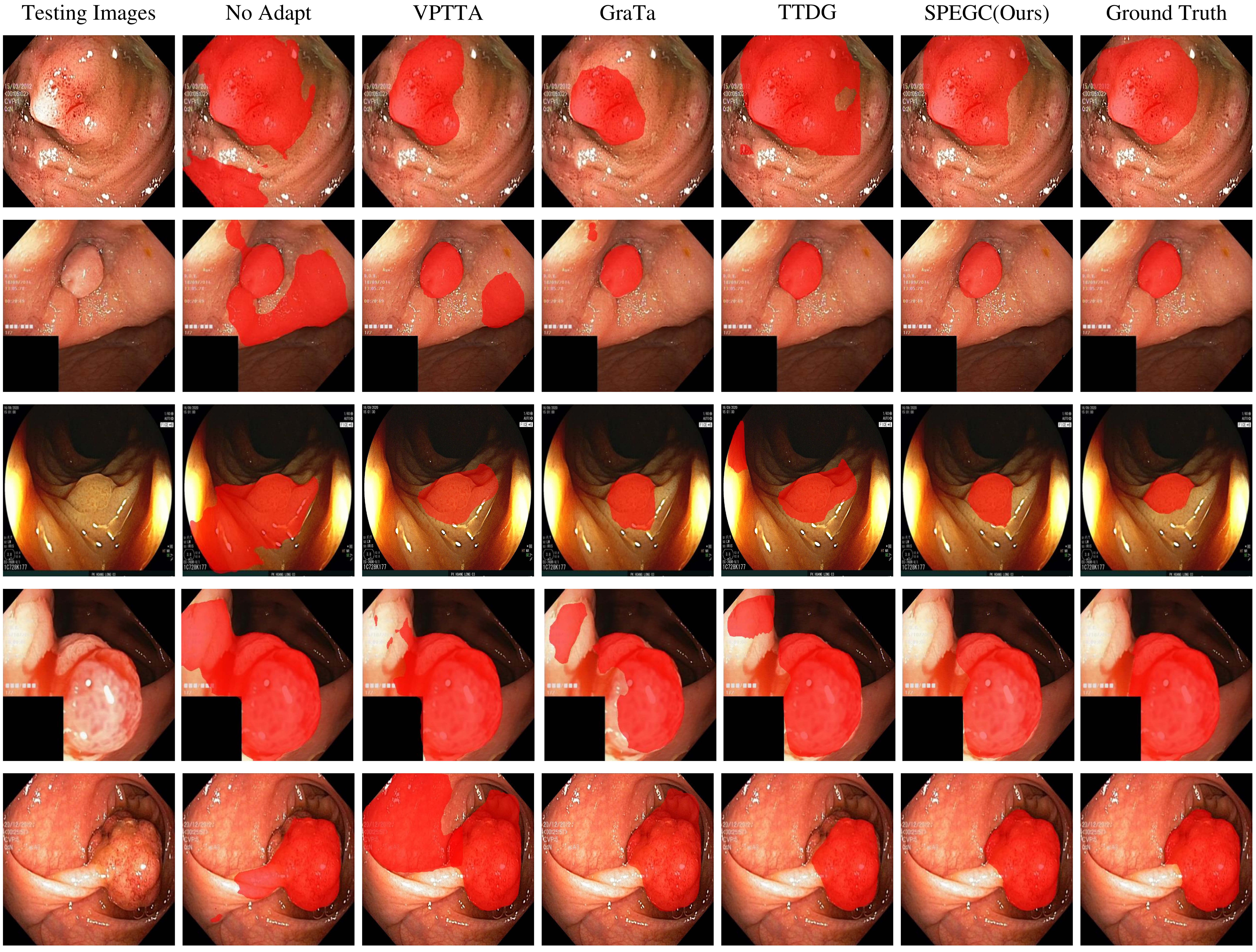}
  
  \caption{
  Visualization comparison of segmentation results for the No Adapt baseline, VPTTA \cite{chen2024each}, GraTa \cite{chen2025gradient}, TTDG \cite{lv2025test}, and SPEGC(Ours) in polyp segmentation.
  }
  
  \label{fig:V_P}
\end{figure*}

In the main paper(\cref{sec:DGCS}), we state that the refined edge similarity matrix, $S^*$, "approximates global consistency." This appendix provides the formal mathematical proof for this claim. We demonstrate that $S^*$ is, by necessity, a differentiable approximation rather than a strict satisfaction of global consistency. This approximation is a direct and necessary consequence of reframing a discrete combinatorial problem into a continuous, end-to-end differentiable optimization landscape.

The objective of our Differentiable Graph Clustering Solver (DGCS) is to extract a refined, continuous affinity structure. While motivated by the fact that a discrete spanning forest with $Z$ components contains $k = V - Z$ edges, DGCS relaxes this discrete partitioning. Instead, it utilizes $k$ as a sparsity budget to perform a differentiable global $k$-edge sparsification via an Optimal Transport (OT) relaxation.

\begin{definition}[Discrete Global Consistency]
We define "true" or "hard" global consistency as the discrete, combinatorial solution to this graph partitioning problem. This solution is a binary assignment vector $y \in \{0, 1\}^E$ (where $E = V^2$ is the total number of possible edges) such that $\sum_{i=1}^E y_i = k$, and $y$ maximizes the total affinity defined by $S'$. Note that while this definition selects $k$ edges to maximize affinity, our OT relaxation does not explicitly enforce the acyclicity required for a true spanning forest, yielding instead a probabilistic soft affinity matrix $S^{*}$.
\end{definition}

This discrete selection process, $y = \text{argmax}(\dots)$, is non-differentiable and thus cannot be used for gradient-based optimization in an end-to-end framework.

To bridge this gap, we reformulate the problem as an optimal transport (OT) task, as introduced in \cref{eq:11,eq:12,eq:13}. The core objective is to find a transport plan $\Gamma \in \mathbb{R}^{E \times 2}$ that maps $E$ edges to two "bins": "reject" or "select", while matching the marginal constraints $r = \mathbf{1}_E$ and $c = [E-k, k]^T$.

Let us first consider the standard, \emph{unregularized} OT problem (i.e., a classic Linear Program):
$$
\Gamma_{\text{hard}}^* = \arg \min_{\Gamma} \langle \Gamma, D \rangle \quad \text{s.t.} \quad \Gamma \mathbf{1}_2 = r, \Gamma^T \mathbf{1}_E = c
$$
where $D$ is the cost matrix defined in \cref{eq:11,eq:12}.

\begin{lemma}[Nature of the Unregularized Solution]
The solution $\Gamma_{\text{hard}}^*$ to this unregularized linear program is guaranteed to be a "hard" assignment. According to the Birkhoff-von Neumann theorem, the vertices of the feasible polytope of $\Gamma$ (the set of doubly-stochastic matrices, or in our case, matrices with fixed marginals) correspond to permutation matrices (or binary assignments). Therefore, the solution $\Gamma_{\text{hard}}^*$ would represent a discrete, binary selection of exactly $k$ edges. The second column, $\Gamma_{\text{hard}, :, 2}^*$, would be the exact binary vector $y$ from Definition 1.
\end{lemma}

While $\Gamma_{\text{hard}}^*$ would satisfy "true" global consistency, solving this linear program is (a) computationally expensive and (b) the solution process is non-differentiable with respect to the input cost matrix $D$.

The critical step in our DGCS is the introduction of the \textbf{entropy regularization} term, $\theta h(\Gamma)$, as shown in \cref{eq:13}:
$$
\Gamma^* = \arg \min_{\Gamma} \langle \Gamma, D \rangle + \theta h(\Gamma)
$$
where $h(\Gamma) = -\sum_{i,j} \Gamma_{i,j} (\log(\Gamma_{i,j}) - 1)$ is the entropy, and $\theta$ is the regularization temperature.

\begin{theorem}[Consequence of Entropy Regularization]
The introduction of the strictly convex entropy term $h(\Gamma)$ (for $\theta > 0$) achieves two goals:
\begin{enumerate}
    \item It makes the objective function strictly convex, guaranteeing a \emph{unique} solution $\Gamma^*$.
    \item It enables the use of the highly efficient, parallel, and—most importantly—\textbf{differentiable} Sinkhorn-Knopp algorithm for finding the solution $\Gamma^*$.
\end{enumerate}
However, this unique solution $\Gamma^*$ is no longer the discrete $\Gamma_{\text{hard}}^*$. The entropy term "softens" the assignment, penalizing sparse (binary) solutions and favoring "diffuse" solutions. The resulting $\Gamma^*$ is a "soft" matrix where entries $\Gamma_{i,j}^* \in (0, 1)$, not $\{0, 1\}$.
\end{theorem}

This $\Gamma^*$ is a differentiable \emph{approximation} of the "hard" linear programming solution $\Gamma_{\text{hard}}^*$. The regularization parameter $\theta$ explicitly controls this trade-off:
\begin{itemize}
    \item As $\theta \to 0$, the problem converges to the "hard" (non-differentiable) linear program, and $\Gamma^* \to \Gamma_{\text{hard}}^*$.
    \item As $\theta \to \infty$, the entropy term dominates, and the solution ignores the cost $D$, becoming $\Gamma_{i,j}^* \propto r_i c_j$.
\end{itemize}
By using a finite, non-zero $\theta$, we explicitly choose to operate in the "soft," approximate regime in order to gain differentiability.

In SPEGC, the final refined edge similarity matrix $S^*$ is constructed by reshaping the second column of this \emph{soft} transport plan: $S^* = \text{reshape}(\Gamma_{:,2}^*)$.

Since $\Gamma^*$ is a "soft" matrix of non-binary values, its second column $\Gamma_{:,2}^*$ is not a binary vector of $k$ ones and $E-k$ zeros. Instead, it is a vector of \textbf{soft probabilities} or "selection likelihoods" for each edge.

Therefore, $S^*$ is not a "hard" adjacency matrix representing a discrete, globally consistent spanning forest. It is, by its very mathematical construction, a \textbf{differentiable approximation} of that ideal structure.

The "approximation" of global consistency is the necessary trade-off for the "differentiability" of the clustering solver. This soft structural representation enables gradient flow via graph consistency loss for end-to-end adaptation.

%% file: sec/A2_algorithm.tex
\section{Algorithm Pipeline}

\cref{alg:spegc} outlines the comprehensive workflow of our proposed SPEGC framework. It details the per-sample continual adaptation procedure, which is structured into three principal stages: (1) Semantic Prompt Feature Enhancement (SPFE), (2) Differentiable Graph Clustering Solver (DGCS), (3) Joint Optimization \& Adaptation.

%% file: sec/A3_additional_visualization.tex
\section{More Experiments}
\subsection{Result Visualization}

To further validate our proposed SPEGC, we provide qualitative comparisons for the continual test-time adaptation (CTTA) stream on both retinal fundus (OD/OC) and polyp segmentation tasks, presented in Figures \cref{fig:V_F} and \cref{fig:V_P}, respectively. Each row visualizes the model's adaptive performance on a distinct, unseen target domain, simulating the challenging real-world scenario of evolving data distributions. The OD/OC segmentation (\cref{fig:V_F}) is inherently difficult, demanding precise delineation of two overlapping anatomical structures (Optic Disc and Cup) often obscured by low contrast and domain-specific artifacts. The polyp segmentation (\cref{fig:V_P}) poses an even greater challenge due to extreme inter-domain variance in lesion morphology, including drastic differences in shape, size, and texture.

As visualized, many competing CTTA methods exhibit significant performance degradation as the domain stream progresses. They suffer from error accumulation or catastrophic forgetting, resulting in noisy predictions, incomplete structures, or a collapse towards source-domain priors. In stark contrast, our SPEGC maintains robust and accurate segmentation. This is primarily attributed to our two-fold mechanism: (1) SPFE injects robust global contextual information, effectively mitigating feature-level noise induced by the domain shift and preserving cross-domain commonalities. (2) DGCS distills a refined, high-order structural representation from these enhanced features. This structure acts as a stable, cluster-level supervisory signal, guiding the model to adapt without compromising core semantic knowledge. As evidenced in the figures, our proposed SPEGC consistently produces precise and structurally coherent segmentation masks that closely align with ground-truth annotations, demonstrating superior stability and adaptation fidelity across diverse and evolving domains, particularly in scenarios where other methods exhibit instability.

\subsection{Comparison experiments under mixed distribution shifts}

To better emulate the complexity of real-world clinical environments---where test data frequently arrive in arbitrarily mixed and continuously evolving streams---we conducted a rigorous evaluation under Mixed Distribution Shifts. In this protocol, a source model trained on a single domain is adapted to a composite target stream, constructed by shuffling samples from all remaining target domains. For reproducibility, the random seed for all data shuffling procedures was fixed at 2026. Quantitative comparisons on the OD/OC and polyp segmentation benchmarks are reported in \cref{tab:mix_ODOC} and \cref{tab:mix_Polyp}, respectively. As evidenced by the results, SPEGC consistently outperforms SOTA methods, achieving the highest average DSC scores of \textbf{83.89\%} and \textbf{78.00\%} across the two tasks. Notably, compared to the runner-up TTDG \cite{lv2025test}, SPEGC exhibits superior stability across varying domains. This highlights the structural robustness and generalization capability of SPEGC in handling complex, online CTTA scenarios.

\subsection{Additional Evaluation Metric: ASSD}
While the Dice Similarity Coefficient (DSC) effectively measures the regional overlap, we additionally introduce the Average Symmetric Surface Distance (ASSD) to further rigorously evaluate the boundary delineations of the segmentation predictions. The ASSD results for the existing 2D tasks (retinal fundus and polyp segmentation) are presented in Tab.~\ref{tab:fundus_assd} and Tab.~\ref{tab:polyp_assd}. For ASSD, we report the metric derived from a single run, where the standard deviation (Std.) is computed across the test images. As observed, SPEGC demonstrates highly competitive performance across virtually all comparisons, indicating that our structural refinement not only improves semantic overlap but also achieves superior boundary consistency.

\begin{table}[h]
    \centering
    \captionsetup{justification=raggedright}
    \caption{Average Symmetric Surface Distance (ASSD, in pixels) for the OD/OC segmentation task. Red and blue indicate the best and second-best results, respectively.}
    \label{tab:fundus_assd}
    \vspace{-6pt}
    \resizebox{\linewidth}{!}{
    \begin{tabular}{c|c|c|c|c|c} 
    \toprule
    Methods & Domain A  & Domain B  & Domain C & Domain D  & Domain E\\
    \midrule \midrule 
    No Adapt & 46.17 & 37.14 & 39.82 & 54.27 &44.72 \\
    \midrule
    SAR & 42.28±30.71 & 35.81±26.93 & 44.59±32.47 & 48.90±27.19 & 47.31±30.72\\
    Domain Adaptor & 40.54±27.54 & 38.17±29.92 & 46.15±31.72 & 51.23±33.02 & 36.28±20.39 \\
    NC-TTT & 38.94±28.19 & \textbf{\textcolor{red}{24.82±20.41}} & 36.54±26.35 & 40.62±28.40 & 30.29±21.79 \\
    VPTTA & 45.28±32.54 & 34.33±26.17 & 46.21±29.86 & 53.81±34.92 & 39.34±30.83 \\ 
    GraTA & 34.62±24.47 & 29.74±23.22 & 35.77±27.58 & 38.04±28.49 & 37.81±26.04 \\
    TTDG & \textbf{\textcolor{red}{27.61±17.33}} & \textbf{\textcolor{blue}{25.16±19.34}} & \textbf{\textcolor{blue}{29.06±21.34}} & \textbf{\textcolor{blue}{33.97±25.71}} & \textbf{\textcolor{red}{23.63±18.49}} \\
    \midrule
    SPEGC (Ours) & \textbf{\textcolor{blue}{29.30±19.67}} & 25.71±16.57 & \textbf{\textcolor{red}{26.33±18.37}} & \textbf{\textcolor{red}{27.64±22.10}} & \textbf{\textcolor{blue}{24.08±20.71}} \\
    \bottomrule
    \end{tabular}
    } 
\end{table}

\begin{table}[h]
    \centering
    \captionsetup{justification=raggedright}
    \caption{Average Symmetric Surface Distance (ASSD, in pixels) for the polyp segmentation task. Red and blue indicate the best and second-best results, respectively.}
    \label{tab:polyp_assd}
    \vspace{-6pt}
    \resizebox{\linewidth}{!}{
    \begin{tabular}{c|c|c|c|c} 
    \toprule
    Methods & Domain A  & Domain B  & Domain C & Domain D \\
    \midrule \midrule 
    No Adapt & 30.49 & 34.36 & 32.19 & 27.02  \\
    \midrule
    SAR & 27.70±12.71 & 35.17±16.97 & 32.04±21.40 & 34.83±15.60 \\
    Domain Adaptor & 23.19±11.94 & 29.76±14.57 & 32.49±19.92 & 33.91±17.42 \\
    NC-TTT & 24.72±14.62 & \textbf{\textcolor{red}{25.57±14.33}} & 33.12±24.63 & 27.43±13.81 \\
    VPTTA & 23.91±12.02 & 27.03±19.72 & 31.99±25.72 & 29.07±16.03 \\
    GraTA & 22.54±16.55 & 30.11±18.06 & \textbf{\textcolor{red}{30.54±23.99}} & 26.39±15.44 \\
    TTDG & \textbf{\textcolor{red}{20.34±13.82}} & 27.69±13.47 & 33.72±21.46 & 27.10±14.52  \\
    \midrule
    SPEGC (Ours) & \textbf{\textcolor{blue}{21.17±12.13}} & \textbf{\textcolor{blue}{26.24±14.09}} & \textbf{\textcolor{blue}{31.27±20.51}} & \textbf{\textcolor{red}{25.71±13.84}} \\
    \bottomrule
    \end{tabular}
    }
\end{table}

\subsection{Extension to 3D Medical Image Segmentation}
To further validate the robustness and applicability of our method across diverse modalities and spatial dimensions, we extend our evaluation to 3D volumetric data using the M\&MS (Multi-Centre, Multi-Vendor \& Multi-Disease Cardiac Image Segmentation) dataset (MRI modality). Following the adaptation paradigm explored in SicTTA\cite{wu2026sictta}, we report both DSC and ASSD for this 3D task. 

As shown in Tab.~\ref{tab:dice_3d_mms} and Tab.~\ref{tab:assd_3d_mms}, SPEGC maintains significant performance gains over the baseline on the 3D M\&MS dataset. Note that for DSC, we conduct five independent runs and report the standard deviation across these runs. A direct quantitative comparison with SicTTA is omitted in these tables due to differences in the underlying backbone architectures; whereas our baseline uniformly employs ResUNet-50, SicTTA utilizes a different backbone network.

\begin{table}[h]
    \centering
    \captionsetup{justification=raggedright}
    \caption{DSC (Mean ± Std.) performance of SPEGC on the 3D M\&MS dataset. Red indicates the best results.}
    \label{tab:dice_3d_mms}
    \vspace{-6pt}
    \resizebox{\linewidth}{!}{
    \begin{tabular}{c|ccc|ccc|ccc|ccc} 
    \toprule
    \multirow{2}{*}{Methods} & \multicolumn{3}{c|}{Domain A} & \multicolumn{3}{c|}{Domain B} & \multicolumn{3}{c|}{Domain C} & \multicolumn{3}{c}{Domain D} \\
    \cmidrule(lr){2-4} \cmidrule(lr){5-7} \cmidrule(lr){8-10} \cmidrule(l){11-13}
    & LV & MYO & RV & LV & MYO & RV & LV & MYO & RV & LV & MYO & RV \\
    \midrule \midrule
    NoAdapt & 83.72 & 70.19 & 72.34 & 76.29 & 69.92 & 68.57 & 77.91 & 68.67 & 65.07 & 78.39 & 68.37 & 64.72 \\
    \midrule
    SPEGC & \textbf{\textcolor{red}{87.64$\pm$1.86}} & \textbf{\textcolor{red}{79.80$\pm$2.72}} & \textbf{\textcolor{red}{75.28$\pm$2.17}} 
    & \textbf{\textcolor{red}{84.16$\pm$1.54}} & \textbf{\textcolor{red}{78.49$\pm$2.63}} & \textbf{\textcolor{red}{75.34$\pm$2.09}} 
    & \textbf{\textcolor{red}{86.49$\pm$1.68}} & \textbf{\textcolor{red}{76.10$\pm$1.82}} & \textbf{\textcolor{red}{70.62$\pm$2.11}} 
    & \textbf{\textcolor{red}{85.17$\pm$1.20}} & \textbf{\textcolor{red}{74.39$\pm$2.17}} & \textbf{\textcolor{red}{69.32$\pm$2.45}} \\
    \bottomrule
    \end{tabular}
}
\end{table}

\begin{table}[h]
    \centering
    \captionsetup{justification=raggedright}
    \caption{ASSD (pixels) (Mean ± Std.) performance of SPEGC on the 3D M\&MS dataset. Red indicates the best results.}
    \label{tab:assd_3d_mms}
    \vspace{-6pt}
    \resizebox{\linewidth}{!}{
    \begin{tabular}{c|ccc|ccc|ccc|ccc} 
    \toprule
    \multirow{2}{*}{Methods} & \multicolumn{3}{c|}{Domain A} & \multicolumn{3}{c|}{Domain B} & \multicolumn{3}{c|}{Domain C} & \multicolumn{3}{c}{Domain D} \\
    \cmidrule(lr){2-4} \cmidrule(lr){5-7} \cmidrule(lr){8-10} \cmidrule(l){11-13} 
    & LV & MYO & RV & LV & MYO & RV & LV & MYO & RV & LV & MYO & RV \\
    \midrule \midrule
    NoAdapt & 4.37 & 4.61 & 4.59 & 4.82 & 5.07 & 5.11 & 4.39 & 5.16 & 5.71 & 5.38 & 4.80 & 4.91 \\
    \midrule
    SPEGC & \textbf{\textcolor{red}{3.76$\pm$3.37}} & \textbf{\textcolor{red}{4.04$\pm$3.92}} & \textbf{\textcolor{red}{4.21$\pm$3.47}} 
    & \textbf{\textcolor{red}{4.17$\pm$3.88}} & \textbf{\textcolor{red}{4.39$\pm$4.01}} & \textbf{\textcolor{red}{4.82$\pm$3.90}} 
    & \textbf{\textcolor{red}{3.26$\pm$3.07}} & \textbf{\textcolor{red}{4.37$\pm$3.52}} & \textbf{\textcolor{red}{5.02$\pm$4.31}} 
    & \textbf{\textcolor{red}{4.14$\pm$2.28}} & \textbf{\textcolor{red}{4.06$\pm$3.16}} & \textbf{\textcolor{red}{4.72$\pm$3.75}} \\
    \bottomrule
    \end{tabular}
    }
\end{table}

\subsection{Sensitivity Analysis of Hyperparameter $\lambda$}
To further investigate the impact of the loss balancing coefficient $\lambda$ (defined in \cref{eq:15} of the main paper) on the adaptation performance, we conduct an additional sensitivity analysis on the OD/OC segmentation task. As presented in Tab.~\ref{tab:lambda_ablation}, the model demonstrates robust stability across a range of values, with the average DSC peaking at $\lambda=0.2$. 

Notably, when $\lambda=0$, the model relies solely on the graph consistency loss $L_G$, yielding sub-optimal results (76.83\%) due to the lack of explicit semantic constraints on the commonality prompt pool. Conversely, performance degrades slightly at higher values (e.g., $\lambda \ge 0.6$) as the clustering loss $L_C$ dominates the optimization, potentially overshadowing the structural guidance provided by $L_G$. Consequently, we adopt $\lambda=0.2$ as the optimal default configuration for all experiments.

\begin{table}[h]
    \centering
    \captionsetup{justification=raggedright}
    \caption{Ablation study of the hyperparameter $\lambda$ on the OD/OC segmentation task (Average DSC). Red indicates the best result.}
    \label{tab:lambda_ablation}
    \vspace{-6pt}
    \resizebox{\linewidth}{!}{
    \begin{tabular}{c|ccccccc} 
    \toprule
    Metric & $\lambda=0$ & $\lambda=0.1$ & $\lambda=0.2$ & $\lambda=0.4$ & $\lambda=0.5$ & $\lambda=0.6$ & $\lambda=0.8$ \\
    \midrule \midrule 
    DSC (\%) & 76.83 & 81.60 & \textbf{\textcolor{red}{84.37}} & 83.79 & 82.14 & 80.42 & 79.59 \\
    \bottomrule
    \end{tabular}
    }
\end{table}